\documentclass{article} 
\usepackage[dvipsnames]{xcolor}
\usepackage{include/iclr2025_conference,times}

\usepackage{amsmath,amsfonts,bm}

\def\eqref#1{equation~\ref{#1}}

\def\1{\bm{1}}

\DeclareMathAlphabet{\mathsfit}{\encodingdefault}{\sfdefault}{m}{sl}
\SetMathAlphabet{\mathsfit}{bold}{\encodingdefault}{\sfdefault}{bx}{n}

\usepackage{hyperref}
\usepackage{url}
\usepackage{titletoc}

\title{Intelligent Go-Explore: Standing on the \\ Shoulders of Giant Foundation Models}

\author{
  Cong Lu\textsuperscript{1,2} \\ \texttt{conglu@cs.ubc.ca} 
  \And Shengran Hu\textsuperscript{1,2} \\ \texttt{srhu@cs.ubc.ca} 
  \And Jeff Clune\textsuperscript{1,2,3} \\ \texttt{jclune@gmail.com}
  \And
  \vspace{-0.7cm}\\ 
  \textsuperscript{1}University of British Columbia\\
  \textsuperscript{2}Vector Institute\\
  \textsuperscript{3}Canada CIFAR AI Chair
}

\iclrfinalcopy 

\usepackage{amsmath}
\usepackage{xspace}
\usepackage{multirow}
\usepackage{booktabs}
\usepackage{color}
\usepackage{colortbl}
\usepackage{cleveref}
\usepackage{graphicx}
\usepackage{algorithm}
\usepackage{algorithmicx}
\usepackage{algpseudocode}
\usepackage{subcaption}
\usepackage{wrapfig}
\usepackage{sidecap}
\usepackage{soul}
\usepackage{enumitem}
\sidecaptionvpos{figure}{t}
\usepackage{multicol}
\usepackage{pifont}
\usepackage[normalem]{ulem}

\newcommand{\xmark}{\ding{55}}

\usepackage[most]{tcolorbox}

\tcbset {
  base/.style={
    arc=0mm, 
    bottomtitle=-0.25mm,
    boxrule=0mm,
    colbacktitle=black!10!white, 
    coltitle=black, 
    fonttitle=\bfseries, 
    left=2.5mm,
    leftrule=1mm,
    right=3.5mm,
    title={#1},
    toptitle=0.25mm,
    breakable,
  }
}

\definecolor{brandblue}{rgb}{0.34, 0.7, 1}
\newtcolorbox{mybox}[1]{
  colframe=brandblue, 
  base={#1}
}

\newcommand{\ouralgolong}{\textsc{Intelligent Go-Explore}\xspace}
\newcommand{\ouralgo}{\textsc{IGE}\xspace}

\definecolor{dodgerblue}{rgb}{0.12, 0.56, 1.0}

\begin{document}

\maketitle

\begin{abstract}
Go-Explore is a powerful family of algorithms designed to solve hard-exploration problems built on the principle of archiving discovered states, and iteratively returning to and exploring from the most promising states.
This approach has led to superhuman performance across a wide variety of challenging problems including Atari games and robotic control, but requires manually designing heuristics to guide exploration (i.e., determine which states to save and explore from, and what actions to consider next), which is time-consuming and infeasible in general.
To resolve this, we propose \ouralgolong (\ouralgo) which greatly extends the scope of the original Go-Explore by replacing these handcrafted heuristics with the intelligence and internalized human notions of interestingness captured by giant pretrained foundation models (FMs).
This provides \ouralgo with a human-like ability to instinctively identify how interesting or promising any new state is (e.g., discovering new objects, locations, or behaviors), even in complex environments where heuristics are hard to define.
Moreover, \ouralgo offers the exciting opportunity to \emph{recognize and capitalize on serendipitous discoveries}---states encountered during exploration that are valuable in terms of exploration, yet where what makes them interesting was not anticipated by the human user.
We evaluate our algorithm on a diverse range of language and vision-based tasks that require search and exploration.
Across these tasks, \ouralgo strongly exceeds classic reinforcement learning and graph search baselines, and also succeeds where prior state-of-the-art FM agents like Reflexion completely fail.
Overall, \ouralgolong combines the tremendous strengths of FMs and the powerful Go-Explore algorithm, opening up a new frontier of research into creating more generally capable agents with impressive exploration capabilities.
All our code is open-sourced at: \url{https://github.com/conglu1997/intelligent-go-explore}.
\end{abstract}

\section{Introduction}
\label{sec:intro}

Foundation models (FMs,~\citet{Bommasani2021FoundationModels, openai2024gpt4,brown2020language,geminiteam2024gemini, touvron2023llama}) trained on giant internet-scale datasets have demonstrated strong general capabilities in reasoning~\citep{wei2022chain} and understanding~\citep{chang2024survey}.
As such, these models have been increasingly employed as autonomous agents~\citep{liu2023agentbench, yao2023react, wang2024survey, yao2023tree, shinn2023reflexion, besta2024graph} in decision-making tasks, showcasing the ability to adapt in-context~\citep{dong2022survey, olsson2022context} to unseen tasks.
Foundation models have also begun to see success in challenging reinforcement learning environments like Doom~\citep{de2024will}, real-world robotic control~\citep{rt1, rt2} and 3D video games~\citep{raad2024scaling, wang2023voyager}.
However, a significant challenge remains: foundation model agents often struggle in environments that require deep exploration over extended time horizons~\citep{liu2023agentbench}.
Overcoming this limitation would enable us to realize their potential as autonomous assistants in more open-ended domains like scientific discovery and innovation~\citep{jiang2023general}.
This paper introduces \ouralgolong (\ouralgo), a novel approach that combines the intelligence of foundation models with the powerful Go-Explore~\citep{first_return, ecoffet2021goexplore} framework to substantially increase the exploration capabilities of FM and reinforcement learning (RL,~\citet{Sutton1998}) agents.

\begin{figure}[t!]
\centering
\vspace{-2mm}
\includegraphics[width=0.9\textwidth, trim={0 0 0 0}, clip]{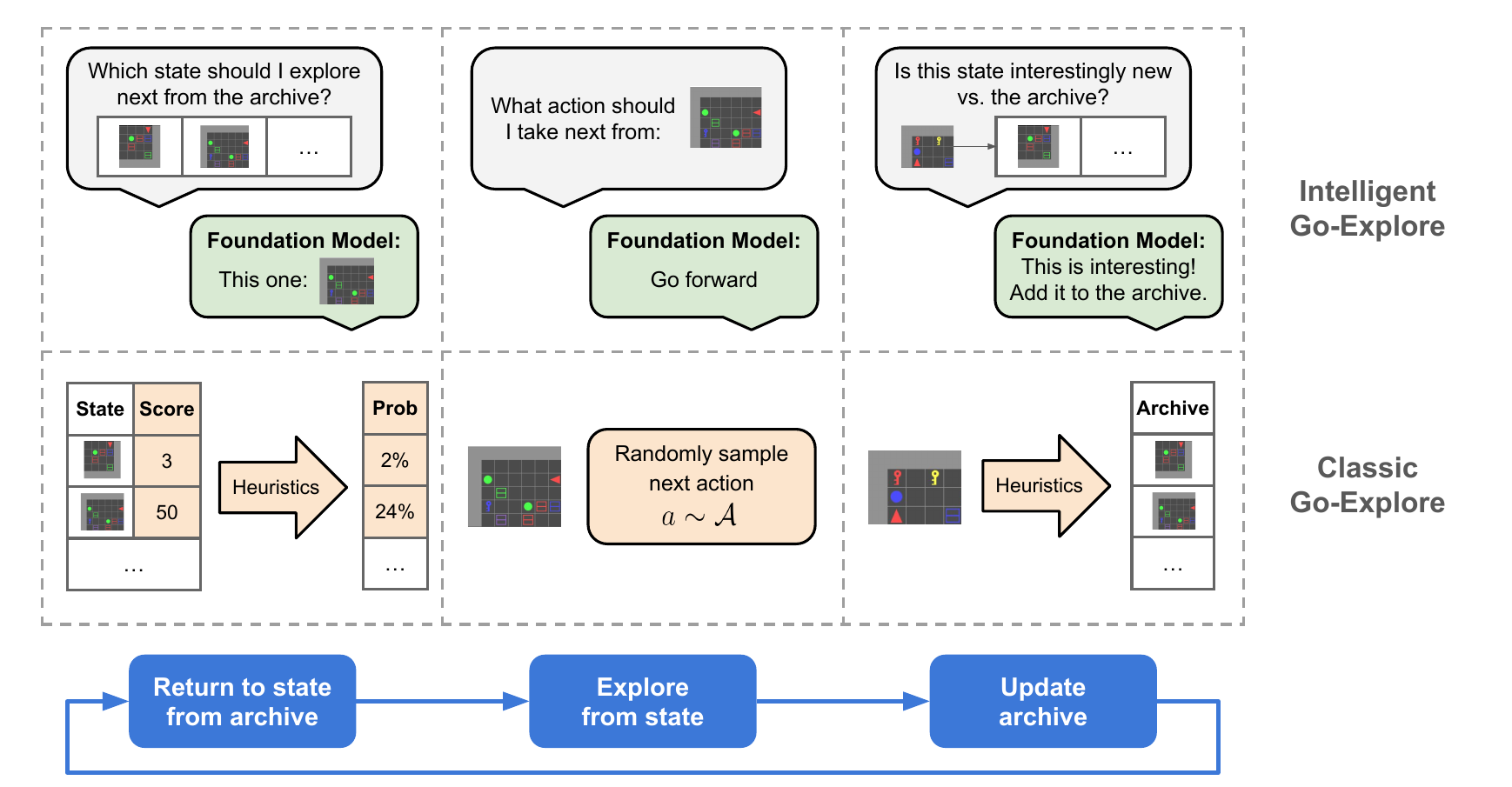}
\caption{
\footnotesize
\ouralgolong (\ouralgo) integrates the intelligence and internalized human notions of interestingness from giant pretrained FMs into all stages of the Go-Explore~\citep{first_return, ecoffet2021goexplore} algorithm, enabling FM agents to robustly explore in complex environments.
\textbf{Bottom:}
Classic Go-Explore solved hard exploration problems by archiving novel discovered states, resetting to promising ones via domain-specific heuristics, and then performing random exploration.
\textbf{Top:}
Our approach, \ouralgolong, enables Go-Explore to tackle virtually any type of problem that is representable in the context of a large language or multimodal model.
Instead of manually defining heuristics, we query the foundation model at all stages, enabling our approach to automatically catch and return to serendipitous discoveries, and harness the power of FM agents to explore.
The environment shown is the BabyAI game used in \Cref{subsec:eval_babyai}.
}
\label{fig:overview}
\vspace{-4mm}
\end{figure}

Go-Explore is a popular family of algorithms in deep RL based on maintaining an archive of ``interestingly new'' discovered states and then iteratively returning to and exploring from the most promising states (see \Cref{fig:overview} for an overview of the three stages).
This framework has led to superhuman performance in a range of hard-exploration problems, including long-horizon Atari games and robotic control.
However, the algorithm's success \emph{largely relies on carefully hand-designed heuristics} at all three stages to guide exploration~\citep{lu_2024_3d, madotto2020exploration}.
For example, in Montezuma's Revenge~\citep{Bellemare_2013}, an Atari game that was the previous grand challenge of exploration in deep RL, \textbf{(1)} saved states in the archive were returned to with probability proportional to factors like the number of times a state has been sampled before, \textbf{(2)} exploration was purely via random action sampling, and \textbf{(3)} the criteria for which states were considered interestingly new enough to be added to the archive depended on domain-specific factors like whether the agent visited a new location, or did so with more keys.
Without this pre-specified knowledge, the quality of discovered trajectories is typically significantly worse~\citep{ecoffet2021goexplore}.

These rigid, domain-specific choices are in stark contrast to human-like exploration of a new game, where players can often intuitively judge the value or interestingness of any particular state~\citep{cooper2014framework}.
More importantly, \emph{it is often impossible to know what is interesting or possible ahead of time} in complex domains.
In the words of Isaac Asimov---\emph{The most exciting phrase to hear in science, the one that heralds new discoveries, is not ``Eureka!'' but ``That's funny.''.}
With this motivation, \ouralgo stands on the shoulders of giant foundation models and uses their intelligence to \textbf{(1)} act as a judge to identify the most promising states to return to and explore from, \textbf{(2)} select the best actions to take from a selected state, and \textbf{(3)} recognize and capitalize on \emph{serendipitous discoveries}---states encountered during exploration that are valuable in terms of exploration, yet where what makes them interesting was not anticipated by the human user.
By leveraging the foundation model's internalized notions of interestingness~\citep{zhang2024omni}, \ouralgo can decide whether a new state is interestingly new enough to be added to the archive as a stepping stone for future exploration (\Cref{fig:overview}, top).

We demonstrate \ouralgo's ability to reliably improve the exploration capabilities of FM agents on a diverse range of language and vision-based tasks that require search and exploration.
These settings include tasks that require \emph{commonsense reasoning, long-term planning and memory, and handling partial observability}.
\ouralgo integrates well with various agent strategies, including few-shot and chain-of-thought-based prompting, demonstrates consistent improvements over baselines across diverse foundation models, and will likely only get better as the capabilities of FMs continue to improve.
While \ouralgo performs strongly all-around, some highlights from our evaluation include: \ouralgo reaches 100\% success rate on Game of 24~\citep{yao2023tree}, a standard mathematical reasoning and search problem, 70.8\% faster than classic graph search.
On the BabyAI domain, \ouralgo enables standard foundation models to succeed with visual observations zero-shot.
Moreover, on the TextWorld~\citep{cote18textworld} Coin Collector domain, \ouralgo is the only algorithm that succeeds in discovering long-horizon optimal solution paths, where prior state-of-the-art FM agent frameworks like Reflexion~\citep{shinn2023reflexion} fail.

\ouralgolong simultaneously empowers foundation model agents to \emph{reliably explore}, and reimagines the scope of Go-Explore to tackle virtually any type of problem, without being limited to hand-designed heuristics.
These abilities will substantially improve our ability to develop more generally capable agents, and increase the range of tasks they can learn how to solve.

\section{Background}
\label{sec:background}
\textbf{Go-Explore for Hard-Exploration Problems.}
Go-Explore~\citep{ecoffet2021goexplore, first_return} is a family of algorithms designed to solve hard-exploration~\citep{ladosz2022exploration} problems based on the principle of remembering and returning reliably to promising states.
The classic setting builds an ``archive'' of novel states it discovers in an environment, where similar states are grouped in a single ``cell''.
These cells are defined by heuristics like having the same visual observation when downsampled to low resolution.
In the beginning, the archive only contains the initial state.
We describe the overall structure of the algorithm in the same order as \Cref{fig:overview} (bottom):
At each iteration, \textbf{(1)} promising states are selected from the archive through domain-specific heuristics, e.g., probabilistically sampling states proportional to their progress through the environment or potential to lead to new states.
The agent returns to that state by resetting using the simulator or via a goal-conditioned policy, and \textbf{(2)} a sequence of random actions is taken to explore from that state.
\textbf{(3)} All discovered states deemed interestingly new by the cell representation heuristics are added to the archive, and the process repeats.
The strength of Go-Explore is due to addressing two critical impediments to exploration: forgetting how to reach previously visited states (detachment) and failing to first return to a state before exploring from it (derailment)~\citep{first_return}.

This approach leads to a collection of high-return trajectories being discovered, which may then be fed into an imitation learning~\citep{hussein2017imitation} algorithm to produce a policy that generalizes and is robust to stochasticity.
We adopt similar assumptions as the original setting, by assuming an agent can return to a previously discovered state by restoring in the simulator (e.g., a reset function in an RL environment).
This assumption may readily be relaxed by training a policy to return to a given state, or in the foundation model case, by simply prompting the model with a past trajectory.

\textbf{Large Language and Multimodal Foundation Models.}
The combination of model scaling and training over internet-scale data has resulted in a wide variety of foundation models~\citep{Bommasani2021FoundationModels} that exhibit generalist capabilities.
In this paper, we consider autoregressive large language models (LLMs,~\citet{brown2020language, openai2024gpt4, touvron2023llama}) which learn to generate text completions by modeling the conditional probability of a new token given the preceding tokens, $p(x_t | x_{<t}; \theta)$.
This framework enables LLMs to not only generate coherent text but crucially also exhibit human-like abilities, including on commonsense knowledge questions~\citep{talmor2019commonsense} and complex reasoning tasks~\citep{wei2022chain}.
These models may also be extended to other input modalities such as images by tokenizing these inputs into the same space as the text~\citep{zhu2023minigpt}.
When prompting an FM with an instruction, the user may decide to do so with no related examples (zero-shot), with a few successful examples in related problems (few-shot,~\citet{brown2020language}), or ask for a chain of reasoning (chain-of-thought,~\citet{wei2022chain}) before responding.

\section{Driving Exploration with Giant Foundation Models}
\label{sec:algo}

In this section, we propose \ouralgolong (\ouralgo) which reimagines the classic Go-Explore algorithm as described in \Cref{sec:background} with the intelligence of giant pretrained foundation models.
Specifically, we introduce FM intelligence for selecting which archived state to return to and explore from, which action to take from each state, and deciding whether a state is interestingly new and should be archived.
\ouralgo's use of foundation models is closely related to FM-as-a-judge~\citep{zheng2023judging}, but instead uses foundation models as proxies of human judgment of exploration choices in an environment rather than the output of generative models.
We illustrate our resultant algorithm at the top of \Cref{fig:overview} and provide full pseudocode in \Cref{alg:main_alg}.

Wherever we query the foundation model, we introduce the overall strategy of Go-Explore alongside a brief description of the current environment in the ``system message'' (high-level directive) displayed below.
The brief descriptions for each environment we evaluate on in \Cref{sec:eval} are listed in \Cref{appsec:envs}.
In the following sections, we detail our prompting techniques at each stage of \ouralgo.
The previous prompt history is visible to the agent, which enables each component of \ouralgo to communicate with each other.
We provide precise details on how we parse responses in \Cref{appsubsec:choice_discussion}.

\begin{mybox}{System Prompt.}
\small
\hl{[Brief Description Of Environment]}\\
You will be prompted to perform systematic exploration in the style of Go-Explore.
An archive will be maintained of interesting states found.
You will be prompted to:\\
- Select a state from the archive that is the most promising, i.e., likely to lead to a solution or more novel states. \\
- Explore from states intelligently, by picking new actions.\\
- For each new state, determine if the state is interestingly new and should be added to the archive.
\end{mybox}

\subsection{Select State From Archive}
\label{subsec:ige_select_state}
The power to easily store and return to promising discovered states is crucial to Go-Explore's ability to reliably solve long-horizon exploration problems.
\ouralgo leverages the foundation model's internalized notions of interestingness~\citep{zhang2024omni} to select the most promising state to return to from the archive (\Cref{fig:overview},~left).
This is far more flexible than classic Go-Explore, which relied on hardcoded hand-crafted heuristics to determine cell sampling probabilities.
An example prompt is shown below.

\begin{wrapfigure}{R}{0.35\textwidth}
\vspace{-8mm}
\begin{mybox}{State Selection Prompt.}
\small
\vspace{-1mm}
Current state archive: \\
\hl{[Discovered states]} \vspace{2mm}\\
Select the most promising state.
\end{mybox}
\vspace{-7mm}
\end{wrapfigure}

Examples of the discovered states are given in \Cref{tab:env_display}.
We assign indices to these states in a list and ask the FM to select a numerical index.
We define a budget of $N_{\text{state}}$ ``state-expansions''.
Each state expansion is followed by a sequence of exploratory actions, which we describe in the next section.

\subsection{Explore From State}
\label{subsec:ige_select_action}
In order to effectively explore from a state selected in the previous section, we leverage the power of foundation model agents~\citep{liu2023agentbench, huang2022language} to choose how to act in an environment.
This vastly improves on the original Go-Explore's use of random action sampling.
One of the key strengths of \ouralgo is that it is \textbf{complementary} to various FM agent reasoning strategies, \emph{including zero-shot, few-shot, or even chain-of-thought-based prompting}~\citep{yao2023react}.
We demonstrate this flexibility in \Cref{sec:eval}.

\begin{wrapfigure}{R}{0.35\textwidth}
\vspace{-6mm}
\begin{mybox}{Action Selection Prompt.}
\small
\hl{[Agent-Specific Prompt]}\\
Current state: \\ \hl{[Current State]} \\
Previously tried actions:\\
\hl{[Previous Action History]} \vspace{2mm} \\
Output the next action.
\end{mybox}
\vspace{-4mm}
\end{wrapfigure}

One point of departure from the classic Go-Explore is that we additionally maintain a state-conditional action history for each archived state, so that \ouralgo can avoid repeating previously tested options.
While this information may already be available in the entire history, this helps \emph{avoid any recency bias that can occur with longer contexts}~\citep{zhao2021calibrate}.
The action history can be easily reiterated in the prompt, or the prompt could display the remaining untested actions.
We define a budget of exploratory actions per state expansion $N_{\text{action}}$, which is typically far shorter than the full horizon of the environment and represents a small number of trial actions.
An example prompt is shown here.

\subsection{Update Archive}
\label{subsec:ige_filter_state}
\ouralgo stores discovered states in an archive, allowing the algorithm to return and explore from those points.
To encourage better exploration, the stored states should be interesting---either promising states relevant to the task that could lead to further stepping stones or novel states that differ significantly from existing ones.
As in other stages of \ouralgo, we leverage the foundation model's internalized notions of interestingness~\citep{zhang2024omni} to evaluate the interestingness of each state.

While the original Go-Explore required extensive domain knowledge to determine interestingness, \ouralgo avoids this requirement and manual labor, critically gaining the ability to recognize and capitalize on serendipitous discoveries that could not have been predicted ahead of time.
In practice, we propose two options to filter discovered states after a sequence of exploratory actions.
The first is to iterate through every new state and ask whether each one is interestingly new and should be added to the archive.
The second is to first add all states and then ask the foundation model to remove the uninteresting states.
We discuss this choice later in \Cref{subsec:eval_textworld}; the second form is preferable in larger environments where there is more need to explicitly deprecate earlier discoveries that have become irrelevant so as not to overload the archive.
An example prompt for the first option is shown below.

\begin{wrapfigure}{R}{0.43\textwidth}
\vspace{-6mm}
\begin{mybox}{Archive Filtering Prompt.}
\small
\vspace{-1mm}
Current state archive: \hl{[State Archive]} \\
New state: \hl{[Current State]} 
\vspace{0.2cm} \\ 
Is this state interestingly new (a novel state that is relevant to the task or could lead to further stepping stones), such that it should be added to the archive?
\end{mybox}
\vspace{-7mm}
\end{wrapfigure}

By default, \ouralgo implements the foundation model at all three stages of Go-Explore, but we rigorously analyze the relative importance of each component in \Cref{sec:analysis}.
Our focus is on the discovery of solutions to hard-exploration problems.
These solutions could potentially be used for downstream reinforcement learning or even improve the foundation model in subsequent tasks through in-context learning, representing exciting avenues for future research.

\begin{table}[t!]
\centering
\caption{
\footnotesize
We show that \ouralgolong can efficiently explore over a diverse set of environments with increasing difficulty.
We showcase \ouralgo on environments with both language and vision-based observations.
For each environment, we provide an example observation, samples from the action space, and the horizon of the task in the environment.
}
\label{tab:env_display}
\resizebox{\textwidth}{!}{
\begin{tabular}{@{}lp{0.3\linewidth}p{0.25\linewidth}p{0.25\linewidth}p{0.4\linewidth}@{}}
\toprule
 & \multicolumn{1}{c}{\textbf{Game of 24}} & \multicolumn{2}{c}{\textbf{BabyAI (Text and Visual)}} & \multicolumn{1}{c}{\textbf{TextWorld}} \\ \midrule
\multirow{3}{*}{\textbf{Problem Type}} & mathematical reasoning and search & \multicolumn{2}{l}{partially observable gridworld with language instructions} & partial observability, long-term planning and memory, and common sense \\ \midrule
\multirow{3}{*}{\textbf{Observation}} & ``Current state: (2 8 8 14)'' & ``Goal: unlock the red door. You see a wall 4 steps forward, You see a yellow box 2 steps left.'' (Text-based) & \centering \raisebox{-0.8\height}{\includegraphics[width=0.6\linewidth]{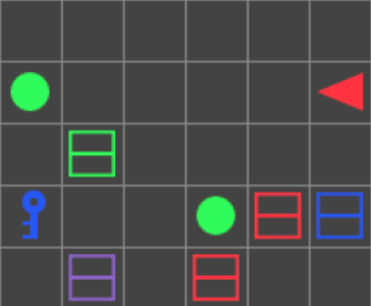}} (Vision-based) & ``You arrive in a pantry... You see a shelf. The shelf is wooden. On the shelf you can see flour...'' \\ \midrule
\textbf{Next Actions} & \begin{tabular}[c]{@{}l@{}}- 2 + 8 = 10 Next: (8 10 14)\\ - 8 / 2 = 4 Next: (4 8 14)\\ - 14 + 2 = 16 Next: (8 8 16)\\ ...\end{tabular} & \multicolumn{2}{l}{\begin{tabular}[c]{@{}l@{}}- turn left\hspace{1.34cm}- pick up\\ - turn right\hspace{1.14cm}- open door\\ - go forward\hspace{1cm}- drop\end{tabular}} & \begin{tabular}[c]{@{}l@{}}- go east\\ - cook potato with oven\\ - unlock door with key\\ ...\end{tabular} \\ \midrule
\textbf{Task Horizon} & \multicolumn{1}{c}{3} & \multicolumn{2}{c}{64 or 128} & \multicolumn{1}{c}{25, 40 or 80} \\ \bottomrule
\end{tabular}
}
\vspace{-4mm}
\end{table}

\section{Empirical Evaluation}
\label{sec:eval}
In this section, we evaluate \ouralgolong across a diverse set of text environments that require search and exploration.
We demonstrate \ouralgo's ability to handle partially observable and complex observation spaces (including both text and visual inputs), discover solutions involving long chains of actions, and effectively improve the ability of FM agents to explore.
For all our experiments, we use GPT-4~\citep{openai2024gpt4}, one of the current SOTA LLMs, as our foundation model.
We compare \ouralgo to random action sampling, a na\"ive LLM baseline, and two SOTA FM agents, ReAct~\citep{yao2023react} and Reflexion~\citep{shinn2023reflexion}.
All methods receive the same environment descriptions, observations, and use the same number of environment steps, ensuring a fair comparison.
Na\"ive LLM simply queries the LLM for an action conditional on the interaction history.
ReAct prompts the agent to output its reasoning before making a decision.
Based on ReAct, Reflexion further conditions the agent on the previous attempted episode, asking the agent to learn from its mistakes.
We provide an overview of our environments in \Cref{tab:env_display}.
Full hyperparameters are detailed in \Cref{appsec:hypers}.

\subsection{Game Of 24}
\label{subsec:eval_game24}

We first demonstrate the effectiveness of \ouralgo in a mathematical reasoning task, Game of 24~\citep{yao2023tree}.
The goal is to perform basic arithmetic operations $(+, -, \times, /)$ starting from 4 numbers to obtain 24.
For example, given input $(4, 9, 10, 13)$, a possible solution could be $(10 - 4) \times (13 - 9) = 24$.
We formulate the problem as an MDP~\citep{Sutton1998}, where actions represent a reduction of two numbers by an arithmetic operation---i.e., the above solution would be represented as the sequence of state transitions $(4, 9, 10, 13) \xrightarrow{10-4=6} (6, 9, 13) \xrightarrow{13-9=4} (6, 4) \xrightarrow{6\times4=24} (24)$.
Therefore, \ouralgo uses the FM to iteratively expand possible solution paths and archive promising ones to return to.
The action space is the range of possible next operations, displayed in the same manner as in \citet{yao2023tree}.

\begin{wrapfigure}{r}{0.40\textwidth}
\vspace{-7mm}
\centering
\includegraphics[width=0.40\textwidth, trim={8 25 8 8},clip]{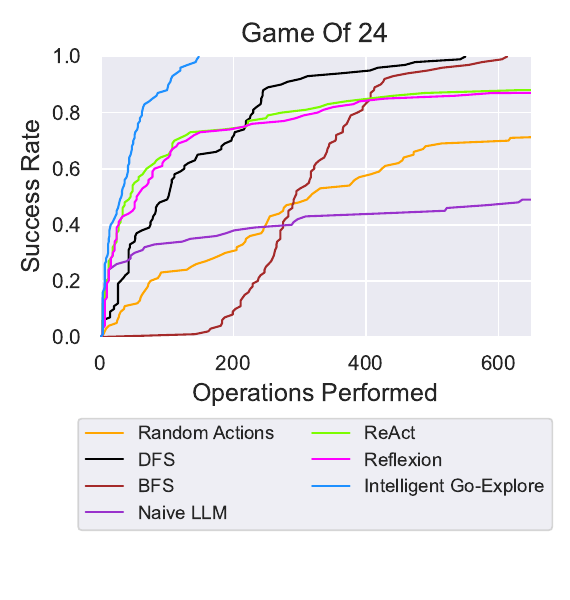}
\vspace{-6mm}
\caption{\footnotesize{
\ouralgo explores the Game of 24 with the intelligence of FMs and reaches 100\% success rate on average 70.8\% faster than DFS, the next best baseline.
\ouralgo completes all problems within 150 environment operations.
Our use of archiving and intelligent action selection allows us to greatly outperform prior LLM agents with an equal number of operations performed.
The success rate is computed over 100 test problems.
}}
\label{fig:game24_sr}
\vspace{-6mm}
\end{wrapfigure}

We evaluate \ouralgo across 100 hard test problems in \Cref{fig:game24_sr}, and additionally include the standard (unweighted) graph search algorithms depth-first search (DFS) and breadth-first search (BFS) as reference.
Since the combinatorial complexity of the problem is at most $\binom{4}{2} \cdot \binom{3}{2} \cdot 4^3 = 1152$, graph search is guaranteed to find a solution within that many actions.
The system prompts for both \ouralgo and the LLM baselines contain \emph{few-shot examples} with correct calculations on different starting numbers.
\ouralgo rapidly reaches 100\% success rate, on average 70.8\% faster than the next best baseline, depth-first search (DFS)---
this improvement is statistically significant ($\chi^2$ test, $p<0.05$) at 150 operations, where \ouralgo has solved all problems.
This success may be attributed to the fact that language models have internalized mathematical intuition and are likely to be able to identify promising pairs like $(6, 4)$ that could easily be multiplied together for a solution.

All LLM agent baselines (na\"ive LLM, ReAct, Reflexion) eventually plateau and even get beaten by the unintelligent DFS.
This highlights the need for diverse action selection, which \ouralgo enables.
A final point of comparison we make is to Tree of Thoughts (ToT,~\citet{yao2023tree}) which achieved 74\% on Game of 24 within their evaluation budget.
We emphasize that our evaluation setting is very different as \ouralgo selects from the list of valid options rather than doing the math in context.
However, we note the key difference to our method is that ToT evaluates and expands multiple reasoning paths following a tree structure, whereas \ouralgo can easily jump around the search space---this is a crucial advantage in more complex environments (like those in the following sections), where it takes many coordinated actions to get from one state to another interesting state.

\subsection{BabyAI Text and Visual Domains}
\label{subsec:eval_babyai}

\begin{figure}[h!]
\vspace{-4mm}
\centering
\begin{subfigure}[b]{0.95\textwidth}
    \centering
    \includegraphics[width=0.99\textwidth]{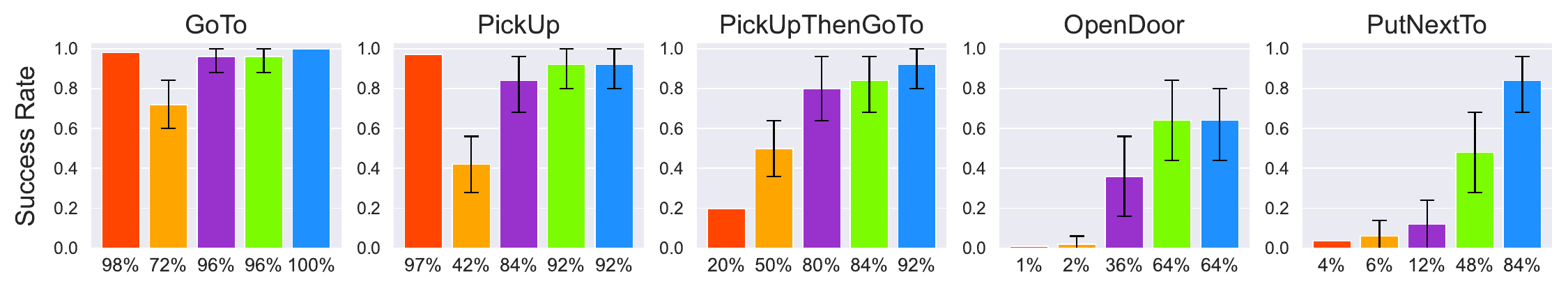}
    \vspace{-2mm}
    \includegraphics[width=0.95\textwidth, trim={10 15 10 15}, clip]{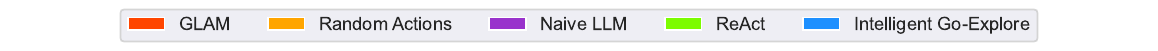}
    \caption{\footnotesize{BabyAI-Text}}
\end{subfigure}
\begin{subfigure}[b]{0.95\textwidth}
    \centering
    \includegraphics[width=0.99\textwidth]{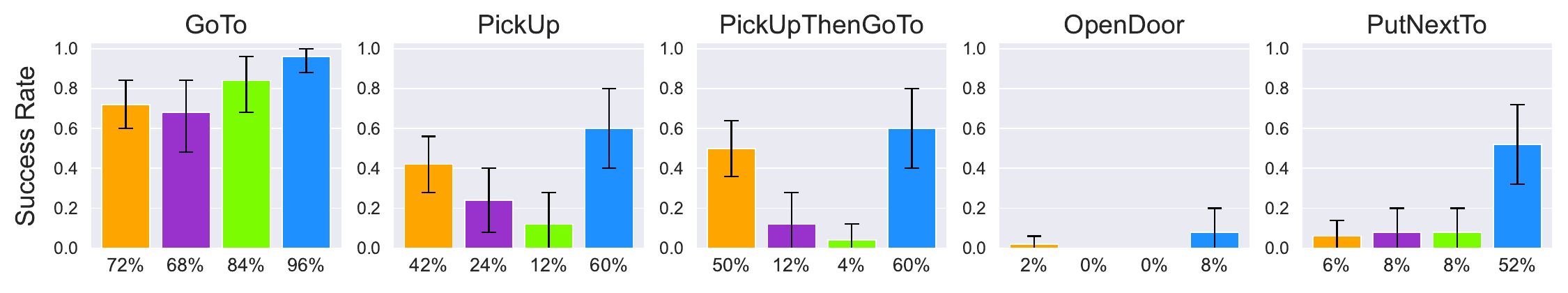}
    \vspace{-2mm}
    \includegraphics[width=0.95\textwidth, trim={10 15 10 15}, clip]{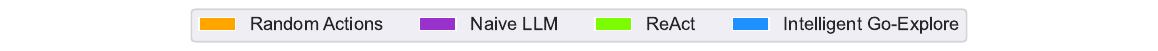}
    \caption{\footnotesize{BabyAI-Visual}}
\end{subfigure}
\vspace{-2mm}
\caption{\footnotesize{
\ouralgo can enable GPT-4o to efficiently find solutions to challenging tasks in the BabyAI text and visual environments. In the text domain, \ouralgo does so with orders of magnitude fewer online steps than prior RL-trained baselines (GLAM,~\citet{pmlr-v202-carta23a}).
Task types are in order of difficulty.
As tasks become more difficult, the performance gap of \ouralgo vs.\ the LLM baselines grows.
We show the mean and 95\% bootstrap confidence interval~\citep{zoubir2007bootstrap} over 25 seeds per environment type.
\emph{Here, and elsewhere, confidence intervals are obtained by bootstrapped resampling 10,000 times.}
}}
\label{fig:babyai_bar}
\vspace{-4mm}
\end{figure}

Next, we show that \ouralgo readily operates across multiple modalities in the BabyAI domains from \citet{pmlr-v202-carta23a}.
The original domain is a procedurally-generated, partially-observable 2D gridworld with text-based observations.
The agent is given a textual goal instruction which could correspond to one or more instructions in a sequence, e.g., ``pick up X and then go to Y''.
As we can see from the observations in \Cref{tab:env_display}, the task is challenging even for humans to complete and requires forming a model of the world from partial observations.
This kind of state observation would make it \emph{hard to define heuristics to determine how good any particular state is}, as in classic Go-Explore.
We additionally extend this to a \textbf{visual domain} by replacing text observations with partially observable image observations.
\ouralgo can naturally handle these by simply replacing text observations with images passed to the multimodal GPT-4o.
The optimal path to a solution may include moving blocking objects as well as finding keys to open doors.
We consider 5 different task families of increasing difficulty: ``go to'', ``pick up'', ``pick up then go to'', ``open door'', and ``put next to'', which are described fully in \Cref{appsubsec:babyai}.

We omit the Reflexion baseline in this environment due to the high cost of querying GPT-4 with 128-step episodes in the context.
Due to the complexity of this environment, we use \emph{chain-of-thought prompting in all three components} of \ouralgo.
This allows the FM to deliberate on the state of the game before making decisions.
\ouralgo can find solutions to these problems with only a tiny budget of 250 environment steps per task (divided into rollouts of 10 exploratory actions each) and visualize the final performance in \Cref{fig:babyai_bar} (top).
In the text variant, \ouralgo and ReAct vastly outperform the prior RL-trained language model approach, GLAM~\citep{pmlr-v202-carta23a}, with orders of magnitude fewer samples (GLAM used 1.5M online steps) and requiring no training whatsoever.
\ouralgo achieves the best or close to the best performance in every task.
The gap between \ouralgo and the second-best method grows with task difficulty, with a statistically significant 36\% improvement ($\chi^2$ test, $p<0.05$) on ``put next to''.
In the visual variant, \ouralgo also strongly beats the visual agent baselines and displays strong performance across the board, enabling GPT-4o to successfully find solutions in a majority of cases.
The results provide strong evidence that \ouralgo extends to any modality representable by a multimodal foundation model.

\subsection{TextWorld}
\label{subsec:eval_textworld}

\begin{figure}[b!]
\vspace{-4mm}
\centering
\includegraphics[width=0.95\textwidth]{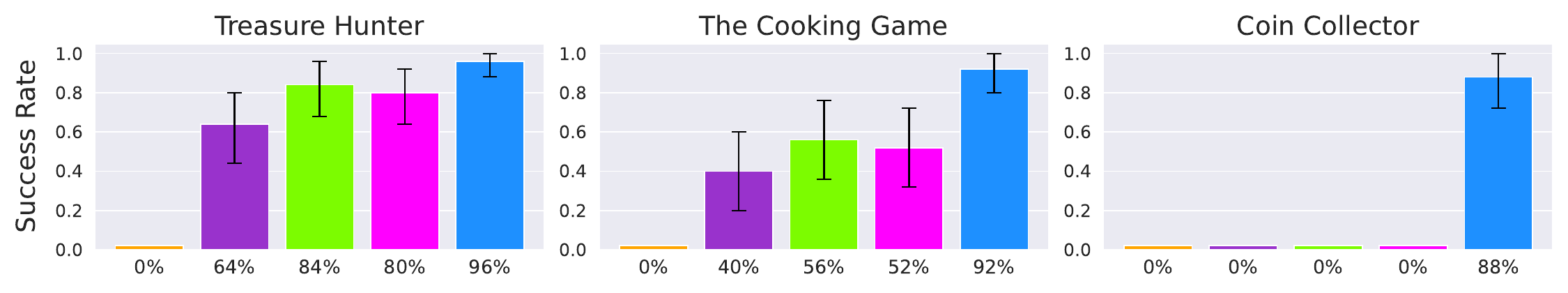}
\vspace{-2mm}
\includegraphics[width=0.9\textwidth, trim={10 15 10 15}, clip]{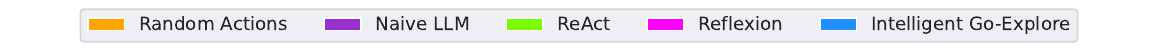}
\vspace{-1mm}
\caption{\footnotesize{
\ouralgo outperforms state-of-the-art FM agents in three challenging text games in TextWorld.
These results illustrate the powerful capabilities of planning, commonsense reasoning, and exploration of \ouralgo (\Cref{subsec:eval_textworld}).
Notably, in the Coin Collector game where hard exploration is required, we observe BFS-like search behavior emerge in \ouralgo, enabling it to find the most efficient solution where all other approaches exhaust the environment horizon.
We show the mean and 95\% bootstrap confidence interval over 25 seeds for each game.
}}
\label{fig:textworld_bar}
\vspace{-4mm}
\end{figure}

Finally, we show \ouralgo's ability to tackle tasks requiring long-horizon memory and planning, exploration, and commonsense in TextWorld~\citep{cote18textworld}, a classic text-based agent benchmark.
We consider three challenging games in TextWorld: Treasure Hunter, The Cooking Game, and Coin Collector.
In each game, the agent needs to complete the task while navigating a maze of different rooms, while only seeing the current room's description in text.
The agent interacts with the world using free-form natural language commands, such as ``go east'' or ``cook potato with oven.''
We set each game to hard difficulty; details on game customizations are provided in \Cref{appsubsec:textworld}.
As in the previous section, we use chain-of-thought prompting in all three components of \ouralgo.
Because the state archive in this environment grows significantly, we implement rejection-based archive filtering, which we describe in \Cref{appsubsec:rejection_archive}.

We present success rates achieved on the three games using \ouralgo and the baselines in \Cref{fig:textworld_bar}.
We observe that \ouralgo outperforms all other baselines, with a statistically significant ($\chi^2$ test, $p<0.05$) gap between \ouralgo and the second-best method in the harder Cooking Game and Coin Collector.
In The Cooking Game, \ouralgo outperforms the next-best agent, ReAct, by a large margin of 36\%, demonstrating \ouralgo's advantage in hard-exploration problems.
In Coin Collector, \ouralgo \emph{is the only method that can find the solution in the maze}, with all other methods completely failing.
Interestingly, we observe that \ouralgo exhibits BFS-like behavior, intelligently selecting rooms with unexplored directions and iteratively removing rooms with exhausted directions.
This results in \ouralgo almost always finding the shortest path to the target, while other methods fail to navigate the maze.

We highlight that Reflexion does not improve over ReAct in all the games we tested.
Although Reflexion should in theory be an improvement over ReAct with the experience from previous attempts, it tends to decrease performance.
We hypothesize that in long-horizon environments, the history becomes too long after the initial episode and prevents Reflexion from effectively utilizing knowledge from the previous episode.
In contrast, \ouralgo uses the FM to iteratively filter interesting states in the archive, which ends up \emph{controlling the context length}.
This helps \ouralgo truly make use of the cumulative knowledge gained through exploration.

\section{Analysis}
\label{sec:analysis}
In this section, we analyze (1) the importance of FM intelligence for each of the three key components of Go-Explore, and (2) how \ouralgo's performance improves as the FM's capabilities increase.
We take a representative sample of environments from the previous section of Game of 24, Put Next To (PN) from BabyAI-Text, and The Cooking Game (CG) from TextWorld.
Hyperparameters are listed in \Cref{appsec:hypers}.

\setlength{\tabcolsep}{15pt}
\begin{table}[t!]
\centering
\footnotesize
\caption{
\footnotesize
We rigorously ablate the design choices in \ouralgo and analyze the importance of incorporating FMs at each stage of the algorithm. Additionally, we compare \ouralgo to Classic Go-Explore~\citep{ecoffet2021goexplore}, which relies on manually defined heuristics at each stage.
Game of 24 performance is taken at 150 environment steps, over 100 evaluation seeds.
BabyAI-Text performance is taken at 250 environment steps, over 25 evaluation seeds.
TextWorld performance is taken at 240 environment steps, over 25 evaluation seeds.
`Standard' is mirrored from \Cref{sec:eval}.
We show the mean and 95\% bootstrap confidence interval for the success rate.
}
\label{tab:ige_ablations}
\vspace{-2mm}
\resizebox{\textwidth}{!}{
\begin{tabular}{lccc}
\toprule
\multicolumn{1}{c}{\multirow{2}{*}{\textbf{Ablation Variants}}} & \multicolumn{3}{c}{\textbf{Success Rate (\%)}}  \\
\multicolumn{1}{c}{}                                              & \textbf{Game of 24} & \textbf{BabyAI (PN)} & \textbf{TextWorld (CG)} \\ \midrule
\ouralgo                                                          & \textbf{100 $\pm$ \phantom{0}0.0}                   & \textbf{84 $\pm$ 14}  & \textbf{92 $\pm$ 10}                   \\
\xmark \ Intelligent action selection                                             & \phantom{0}68 $\pm$ \phantom{0}9.0                   & 24 $\pm$ 16    & \phantom{0}0 $\pm$ \phantom{0}0                \\
\xmark \ Intelligent state selection                                              & \phantom{0}96 $\pm$ \phantom{0}3.5                   & 48 $\pm$ 20        & 76 $\pm$ 16             \\
\xmark \ Intelligent archive filtering                                              & \phantom{0}93 $\pm$ \phantom{0}5.0          & 64 $\pm$ 20     & 64 $\pm$ 20       \\
\xmark \ All 3 above                                                             & \phantom{0}61 $\pm$ \phantom{0}9.5           & \phantom{0}4 $\pm$ \phantom{0}6  & \phantom{0}0 $\pm$ \phantom{0}0           \\
\xmark \ State-conditional action history                                             & \phantom{0}33 $\pm$ \phantom{0}9.0                   & 72 $\pm$ 16      & 72 $\pm$ 16       \\ 
\midrule
Classic Go-Explore (Visitation Frequency) & \phantom{0}38 $\pm$ 22.0 & \phantom{0}4 $\pm$ \phantom{0}6 & \phantom{0}0 $\pm$ \phantom{0}0      \\ 
\bottomrule
\end{tabular}
}
\vspace{-4mm}
\end{table}

\textbf{How Important is Foundation Model Intelligence at Each Step?}
First, we analyze the impact of FM intelligence on each component of \ouralgolong.
We ablate replacing state and action selection with uniform random sampling, archive filtering with saving everything to the archive, and not maintaining a state-conditional action history.
We use these unintelligent choices, as \emph{it would be very time-consuming to attempt to design the right heuristics} based on the rich text observations in \Cref{tab:env_display}.
In \Cref{tab:ige_ablations}, we observe that where the intelligence of FMs is more valuable varies by environment.
Since the environment horizon is only 3 in the Game of 24, the most important factor is ensuring that the actions tried are diverse and intelligently selected.
This hypothesis is confirmed: the largest performance drops occur when removing either FM action selection or the action history.
Different IGE components are most helpful in both of the longer-horizon BabyAI-Text and TextWorld environments: intelligent state selection and archive filtering make a big impact, showcasing the strength of enabling \ouralgo to return to promising discovered states.
There are smaller performance drops when removing the action history; likely because in larger environments, many more unique states are discovered, so there is less gain from preventing taking the same actions from frequently returned to states.
In both environments, we also observe a drastic decrease when switching to random actions, as in classic Go-Explore.
This underscores the substantial benefits \ouralgo provides in harnessing FMs for action selection.

Next, we elucidate the need for intelligent archive filtering across all our environments.
Not only does archive filtering improve performance, but it also drastically reduces the number of uninteresting states in the archive.
For instance, in BabyAI-Text, we observe the archive becoming around $8\times$ larger without filtering.
These metrics demonstrate \ouralgo's innate ability to capture promising discoveries as they occur and focus attention on them, without the need for any manual heuristics.
Detailed analysis and results are provided in \Cref{appsec:archive_size}.
Furthermore, we compare \ouralgo with classic Go-Explore (GE,~\citet{ecoffet2021goexplore}), which adopts fixed pre-defined, hand-designed, open-loop heuristics like resetting to states with probability inversely proportional to the state-visitation count.
Across all 3 domains, \ouralgo significantly outperforms classic GE, which shows that adopting FMs instead of manually designed heuristics not only saves the effort required for domain-specific design but also enhances the algorithm's exploration by leveraging the human notion of interestingness captured by FMs (\Cref{tab:ige_ablations}, bottom).
We particularly emphasize that the intelligence of FMs enables \ouralgo to rapidly solve environments like BabyAI that have challenged the RL community in an exceptionally short number of 250 timesteps.
This is in contrast to the classic GE paradigm, which used millions or even billions of environment steps~\citep{ecoffet2021goexplore}.

\textbf{What is the Effect of Foundation Model Choice?}
We further analyze the dependence of \ouralgo on the capabilities of the underlying foundation model.
We evaluated \ouralgo using other foundation models, including Claude Sonnet 3.5~\citep{claude3} and the open-source Llama-3 400B~\citep{llama3}, on the Game of 24.
The results, presented in \Cref{tab:ige_model_comparison}, show that \ouralgo maintains superior performance over the baselines across different foundation models, consistently achieving higher success rates.
For instance, even when using the weaker Llama-3 400B, \ouralgo achieves a success rate of $98\%$, significantly outperforming the best Llama-based baseline.
Notably, the performance gains of \ouralgo are statistically significant (for all baselines with Llama-3 400B and for all except Reflexion on Claude Sonnet 3.5, $p < 0.05$).
This demonstrates that \ouralgo is not dependent on a specific model and can be expected to perform well with other, possibly future, foundation models with higher capabilities.
Additional experiments on prompt robustness are provided in \Cref{appsubsec:prompt_robustness}.

\setlength{\tabcolsep}{18.5pt}
\begin{table}[h!]
\footnotesize
\centering
\vspace{-2mm}
\caption{
\footnotesize
\ouralgo consistently outperforms other LLM agent baselines with a diverse range of foundation models on the Game of 24.
We use the same evaluation setting as \Cref{tab:ige_ablations} and show the mean and 95\% bootstrap confidence interval for the success rate (\%).
}
\vspace{-2mm}
\label{tab:ige_model_comparison}
\begin{tabular}{lcccc}
\toprule
\textbf{Model}          & \textbf{Na\"ive LLM}       & \textbf{ReAct}           & \textbf{Reflexion}      & \textbf{\ouralgo}                   \\ \midrule
GPT-4            & $70 \pm 12$     & $82 \pm 11$     & $83 \pm 10$    & $\mathbf{100 \pm 0}$  \\
Llama-3 400B     & $44 \pm 14$     & $68 \pm 13$     & $54 \pm 14$    & $\mathbf{\hphantom{0}98 \pm 3}$  \\ 
Claude Sonnet 3.5       & $24 \pm 12$     & $58 \pm 14$     & $80 \pm 11$    & $\mathbf{\hphantom{0}86 \pm 9}$  \\ \bottomrule
\end{tabular}
\vspace{-4mm}
\end{table}

\section{Related Work}
\label{sec:related_work}

\emph{Due to space constraints, extra related work discussions are in \Cref{appsec:more_related_work}.}

\textbf{FM Agents.}
One of the key strengths of \ouralgo is that it is agnostic to the precise agent formulation and thus strictly additive on top of a wide variety of strategies.
A common strategy is chain-of-thought-based methods~\citep{yao2023react, hu2023ThoughtCloning}, which prompts the FM to output a set of reasoning steps before the answer.
We integrate this into the FM guidance in our experiments in \Cref{subsec:eval_babyai,subsec:eval_textworld}.
Reflexion~\citep{shinn2023reflexion} enables an agent to improve over multiple episodes by asking it to reflect on the previous attempted episode and learn from its mistakes.
However, we show this can break down in tasks with long horizons, while \ouralgo proposes a more efficient way to filter out the vast majority of uninteresting interactions.
Another set of agent frameworks that are related to the idea of exploring diverse solution paths via state-connectivity is Tree of Thoughts (ToT,~\citet{yao2023tree}) and Graph of Thoughts (GoT,~\citet{Besta_2024}).
The primary difference between methods like ToT and \ouralgo is that ToT builds up a tree of abstract thoughts generated by the language model and could be hallucinated, whereas \ouralgo's archive is grounded in real states directly copied from an RL environment.
Although ToT's thoughts are suitable for the types of problems it investigates, these abstract plans could easily break down in RL environments with long horizons.
In contrast, \ouralgo scales gracefully with long horizons by archiving real states and returning to the state for exploration.
Additionally, ToT is based on DFS and BFS over the tree and is thus limited to these search strategies over the problem.
On the other hand, \ouralgo can intelligently select hybrid strategies, for example, going broad at the start (more BFS-like) and then honing down on a promising path (more DFS-like) later on.
This is particularly important for long-horizon tasks with larger state spaces, as we show in \Cref{subsec:eval_babyai,subsec:eval_textworld}.

Closely related to exploration, FM agents have also begun to see use in search-based tasks.
Stream of Search~\citep{gandhi2024stream} considers a similar mathematical reasoning task to the Game of 24 and seeks to initially clone the actions of graph search algorithms, then use RL to self-improve.
In contrast, \ouralgo already greatly outperforms classic graph search, and an exciting future direction could be to first clone the exploratory behavior of \ouralgo and then use that as a basis for self-improvement.
\citet{lehnert2024a} analogously train a language model to mimic the A$^\ast$ algorithm.
Finally, \citet{krishnamurthy2024large} also consider bootstrapping exploration with an externally summarized action-history in bandit problems; our focus is more on the detection of interesting states.

\textbf{Go-Explore.}
The original Go-Explore~\citep{first_return,ecoffet2021goexplore} framework enabled superhuman performance in a variety of hard-exploration problems, including applications as diverse as automated game testing~\citep{lu_2024_3d}.
\citet{gallouédec2023cellfree} propose Latent Go-Explore which similarly aims to address the difficulty of designing exploration heuristics by automatically learning a latent representation and sampling states with a low latent density.
However, this requires periodic retraining and could easily miss out on rare discoveries.
HuGE~\citep{NEURIPS2023_c7c7cf10} guides Go-Explore with humans in the loop by asking for pair-wise feedback on which goal to select.
On the other hand, we replace humans with intelligent FM guidance at all components of Go-Explore.

\section{Conclusion and Limitations}
\label{sec:conclusion}
In this paper, we demonstrate a new approach to robust exploration in complex environments, \ouralgolong, reimagining Go-Explore in the era of giant foundation models.
We show that \ouralgo can drive exploration for a diverse set of FM agents, including few-shot and chain-of-thought prompting, across a variety of modalities including challenging text and vision-based games.
Extending \ouralgo to even more modalities could unlock applications as wide as scientific discovery in synthetic biology (designing novel drugs or proteins) or material science.
Recent advancements in multimodal foundation models, such as RT-2~\citep{rt2} and RFM-1~\citep{covariant_2024_rfm}, have shown the potential of FMs to handle various modalities, including text, images, videos, and (continuous) numerical sensor readings.
Since these models have already tokenized the state-action spaces of general environments and shown the ability of the model to generate actions for any state, it should be possible to then ask the FM to judge the interestingness of any given state compared to prior states observed, which is the only additional requirement of \ouralgo.
A further interesting domain is the (hitherto unsolved by intelligent agents) vast dungeon crawler, NetHack~\citep{kuttler2020nethack}.
\citet{kuttler2020nethack} noted that classic Go-Explore's heuristics ``will likely not work for large symbolic and procedurally generated environments.''
\ouralgo represents a sharp departure from these limitations by replacing hard-coded and inflexible exploration heuristics with the dynamic intelligence of giant foundation models.

There remain exciting opportunities to improve \ouralgo's capabilities to explore vast state spaces.
For example, we currently recall and compare against the entire archive whenever we discover a new state.
This could be made much more efficient by using techniques like retrieval augmented generation~\citep{NEURIPS2020_6b493230} and only comparing to the closest previously discovered states.
As we consider \ouralgo for real-world settings, we should take steps to ensure the responsible deployment of FMs~\citep{Bommasani2021FoundationModels}.
Our approach opens up the road to \textbf{safe and interpretable exploration}: through careful prompt engineering or techniques like constitutional AI~\citep{bai2022constitutional}, we could steer the agent away from unsafe behaviors.
Furthermore, if we ask or train the FM to explain its choices in each part of \ouralgo, we could gain insight into its rationale for exploring particular paths through an environment~\citep{wei2022chain, hu2023ThoughtCloning}; improving safety, interpretability, and perhaps one day even our own understanding of how best to explore.

\subsubsection*{Acknowledgments}
This work was supported by the Vector Institute, the Canada CIFAR AI Chairs program, grants from Schmidt Futures and Open Philanthropy, an NSERC Discovery Grant, and a generous donation from Rafael Cosman.
We thank Aaron Dharna, Ben Norman, and Jenny Zhang from our lab at the University of British Columbia for insightful discussions and feedback on early drafts of this work.

\bibliography{references}

\begin{thebibliography}{61}
\providecommand{\natexlab}[1]{#1}
\providecommand{\url}[1]{\texttt{#1}}
\expandafter\ifx\csname urlstyle\endcsname\relax
  \providecommand{\doi}[1]{doi: #1}\else
  \providecommand{\doi}{doi: \begingroup \urlstyle{rm}\Url}\fi

\bibitem[Anthropic(2024)]{claude3}
Anthropic.
\newblock The claude 3 model family: Opus, sonnet, haiku, 2024.
\newblock URL \url{https://www-cdn.anthropic.com/de8ba9b01c9ab7cbabf5c33b80b7bbc618857627/Model_Card_Claude_3.pdf}.

\bibitem[Bai et~al.(2022)Bai, Kadavath, Kundu, Askell, Kernion, Jones, Chen, Goldie, Mirhoseini, McKinnon, Chen, Olsson, Olah, Hernandez, Drain, Ganguli, Li, Tran-Johnson, Perez, Kerr, Mueller, Ladish, Landau, Ndousse, Lukosuite, Lovitt, Sellitto, Elhage, Schiefer, Mercado, DasSarma, Lasenby, Larson, Ringer, Johnston, Kravec, Showk, Fort, Lanham, Telleen-Lawton, Conerly, Henighan, Hume, Bowman, Hatfield-Dodds, Mann, Amodei, Joseph, McCandlish, Brown, and Kaplan]{bai2022constitutional}
Yuntao Bai, Saurav Kadavath, Sandipan Kundu, Amanda Askell, Jackson Kernion, Andy Jones, Anna Chen, Anna Goldie, Azalia Mirhoseini, Cameron McKinnon, Carol Chen, Catherine Olsson, Christopher Olah, Danny Hernandez, Dawn Drain, Deep Ganguli, Dustin Li, Eli Tran-Johnson, Ethan Perez, Jamie Kerr, Jared Mueller, Jeffrey Ladish, Joshua Landau, Kamal Ndousse, Kamile Lukosuite, Liane Lovitt, Michael Sellitto, Nelson Elhage, Nicholas Schiefer, Noemi Mercado, Nova DasSarma, Robert Lasenby, Robin Larson, Sam Ringer, Scott Johnston, Shauna Kravec, Sheer~El Showk, Stanislav Fort, Tamera Lanham, Timothy Telleen-Lawton, Tom Conerly, Tom Henighan, Tristan Hume, Samuel~R. Bowman, Zac Hatfield-Dodds, Ben Mann, Dario Amodei, Nicholas Joseph, Sam McCandlish, Tom Brown, and Jared Kaplan.
\newblock Constitutional ai: Harmlessness from ai feedback, 2022.

\bibitem[Bellemare et~al.(2013)Bellemare, Naddaf, Veness, and Bowling]{Bellemare_2013}
M.~G. Bellemare, Y.~Naddaf, J.~Veness, and M.~Bowling.
\newblock The arcade learning environment: An evaluation platform for general agents.
\newblock \emph{Journal of Artificial Intelligence Research}, 47:\penalty0 253–279, June 2013.
\newblock ISSN 1076-9757.
\newblock \doi{10.1613/jair.3912}.
\newblock URL \url{http://dx.doi.org/10.1613/jair.3912}.

\bibitem[Bengio et~al.(2023)Bengio, Lahlou, Deleu, Hu, Tiwari, and Bengio]{bengio2023gflownetfoundations}
Yoshua Bengio, Salem Lahlou, Tristan Deleu, Edward~J. Hu, Mo~Tiwari, and Emmanuel Bengio.
\newblock Gflownet foundations, 2023.
\newblock URL \url{https://arxiv.org/abs/2111.09266}.

\bibitem[Besta et~al.(2024{\natexlab{a}})Besta, Blach, Kubicek, Gerstenberger, Podstawski, Gianinazzi, Gajda, Lehmann, Niewiadomski, Nyczyk, and Hoefler]{Besta_2024}
Maciej Besta, Nils Blach, Ales Kubicek, Robert Gerstenberger, Michal Podstawski, Lukas Gianinazzi, Joanna Gajda, Tomasz Lehmann, Hubert Niewiadomski, Piotr Nyczyk, and Torsten Hoefler.
\newblock Graph of thoughts: Solving elaborate problems with large language models.
\newblock \emph{Proceedings of the AAAI Conference on Artificial Intelligence}, 38\penalty0 (16):\penalty0 17682–17690, March 2024{\natexlab{a}}.
\newblock ISSN 2159-5399.
\newblock \doi{10.1609/aaai.v38i16.29720}.
\newblock URL \url{http://dx.doi.org/10.1609/aaai.v38i16.29720}.

\bibitem[Besta et~al.(2024{\natexlab{b}})Besta, Blach, Kubicek, Gerstenberger, Podstawski, Gianinazzi, Gajda, Lehmann, Niewiadomski, Nyczyk, et~al.]{besta2024graph}
Maciej Besta, Nils Blach, Ales Kubicek, Robert Gerstenberger, Michal Podstawski, Lukas Gianinazzi, Joanna Gajda, Tomasz Lehmann, Hubert Niewiadomski, Piotr Nyczyk, et~al.
\newblock Graph of thoughts: Solving elaborate problems with large language models.
\newblock In \emph{Proceedings of the AAAI Conference on Artificial Intelligence}, 2024{\natexlab{b}}.

\bibitem[Bommasani et~al.(2021)Bommasani, Hudson, Adeli, Altman, Arora, von Arx, and et~al.]{Bommasani2021FoundationModels}
Rishi Bommasani, Drew~A. Hudson, Ehsan Adeli, Russ Altman, Simran Arora, Sydney von Arx, and Michael S.~Bernstein et~al.
\newblock On the opportunities and risks of foundation models.
\newblock \emph{ArXiv}, 2021.
\newblock URL \url{https://crfm.stanford.edu/assets/report.pdf}.

\bibitem[Bradley et~al.(2023)Bradley, Dai, Teufel, Zhang, Oostermeijer, Bellagente, Clune, Stanley, Schott, and Lehman]{bradley2023qualitydiversity}
Herbie Bradley, Andrew Dai, Hannah Teufel, Jenny Zhang, Koen Oostermeijer, Marco Bellagente, Jeff Clune, Kenneth Stanley, Grégory Schott, and Joel Lehman.
\newblock Quality-diversity through ai feedback, 2023.

\bibitem[Brohan et~al.(2022)Brohan, Brown, Carbajal, Chebotar, Dabis, Finn, Gopalakrishnan, Hausman, Herzog, Hsu, Ibarz, Ichter, Irpan, Jackson, Jesmonth, Joshi, Julian, Kalashnikov, Kuang, Leal, Lee, Levine, Lu, Malla, Manjunath, Mordatch, Nachum, Parada, Peralta, Perez, Pertsch, Quiambao, Rao, Ryoo, Salazar, Sanketi, Sayed, Singh, Sontakke, Stone, Tan, Tran, Vanhoucke, Vega, Vuong, Xia, Xiao, Xu, Xu, Yu, and Zitkovich]{rt1}
Anthony Brohan, Noah Brown, Justice Carbajal, Yevgen Chebotar, Joseph Dabis, Chelsea Finn, Keerthana Gopalakrishnan, Karol Hausman, Alex Herzog, Jasmine Hsu, Julian Ibarz, Brian Ichter, Alex Irpan, Tomas Jackson, Sally Jesmonth, Nikhil Joshi, Ryan Julian, Dmitry Kalashnikov, Yuheng Kuang, Isabel Leal, Kuang-Huei Lee, Sergey Levine, Yao Lu, Utsav Malla, Deeksha Manjunath, Igor Mordatch, Ofir Nachum, Carolina Parada, Jodilyn Peralta, Emily Perez, Karl Pertsch, Jornell Quiambao, Kanishka Rao, Michael Ryoo, Grecia Salazar, Pannag Sanketi, Kevin Sayed, Jaspiar Singh, Sumedh Sontakke, Austin Stone, Clayton Tan, Huong Tran, Vincent Vanhoucke, Steve Vega, Quan Vuong, Fei Xia, Ted Xiao, Peng Xu, Sichun Xu, Tianhe Yu, and Brianna Zitkovich.
\newblock Rt-1: Robotics transformer for real-world control at scale.
\newblock In \emph{arXiv preprint arXiv:2212.06817}, 2022.

\bibitem[Brohan et~al.(2023)Brohan, Brown, Carbajal, Chebotar, Chen, Choromanski, Ding, Driess, Dubey, Finn, Florence, Fu, Arenas, Gopalakrishnan, Han, Hausman, Herzog, Hsu, Ichter, Irpan, Joshi, Julian, Kalashnikov, Kuang, Leal, Lee, Lee, Levine, Lu, Michalewski, Mordatch, Pertsch, Rao, Reymann, Ryoo, Salazar, Sanketi, Sermanet, Singh, Singh, Soricut, Tran, Vanhoucke, Vuong, Wahid, Welker, Wohlhart, Wu, Xia, Xiao, Xu, Xu, Yu, and Zitkovich]{rt2}
Anthony Brohan, Noah Brown, Justice Carbajal, Yevgen Chebotar, Xi~Chen, Krzysztof Choromanski, Tianli Ding, Danny Driess, Avinava Dubey, Chelsea Finn, Pete Florence, Chuyuan Fu, Montse~Gonzalez Arenas, Keerthana Gopalakrishnan, Kehang Han, Karol Hausman, Alexander Herzog, Jasmine Hsu, Brian Ichter, Alex Irpan, Nikhil Joshi, Ryan Julian, Dmitry Kalashnikov, Yuheng Kuang, Isabel Leal, Lisa Lee, Tsang-Wei~Edward Lee, Sergey Levine, Yao Lu, Henryk Michalewski, Igor Mordatch, Karl Pertsch, Kanishka Rao, Krista Reymann, Michael Ryoo, Grecia Salazar, Pannag Sanketi, Pierre Sermanet, Jaspiar Singh, Anikait Singh, Radu Soricut, Huong Tran, Vincent Vanhoucke, Quan Vuong, Ayzaan Wahid, Stefan Welker, Paul Wohlhart, Jialin Wu, Fei Xia, Ted Xiao, Peng Xu, Sichun Xu, Tianhe Yu, and Brianna Zitkovich.
\newblock Rt-2: Vision-language-action models transfer web knowledge to robotic control, 2023.
\newblock URL \url{https://arxiv.org/abs/2307.15818}.

\bibitem[Brown et~al.(2020)Brown, Mann, Ryder, Subbiah, Kaplan, Dhariwal, Neelakantan, Shyam, Sastry, Askell, Agarwal, Herbert-Voss, Krueger, Henighan, Child, Ramesh, Ziegler, Wu, Winter, Hesse, Chen, Sigler, Litwin, Gray, Chess, Clark, Berner, McCandlish, Radford, Sutskever, and Amodei]{brown2020language}
Tom~B. Brown, Benjamin Mann, Nick Ryder, Melanie Subbiah, Jared Kaplan, Prafulla Dhariwal, Arvind Neelakantan, Pranav Shyam, Girish Sastry, Amanda Askell, Sandhini Agarwal, Ariel Herbert-Voss, Gretchen Krueger, Tom Henighan, Rewon Child, Aditya Ramesh, Daniel~M. Ziegler, Jeffrey Wu, Clemens Winter, Christopher Hesse, Mark Chen, Eric Sigler, Mateusz Litwin, Scott Gray, Benjamin Chess, Jack Clark, Christopher Berner, Sam McCandlish, Alec Radford, Ilya Sutskever, and Dario Amodei.
\newblock Language models are few-shot learners, 2020.

\bibitem[Carta et~al.(2023)Carta, Romac, Wolf, Lamprier, Sigaud, and Oudeyer]{pmlr-v202-carta23a}
Thomas Carta, Cl\'{e}ment Romac, Thomas Wolf, Sylvain Lamprier, Olivier Sigaud, and Pierre-Yves Oudeyer.
\newblock Grounding large language models in interactive environments with online reinforcement learning.
\newblock In Andreas Krause, Emma Brunskill, Kyunghyun Cho, Barbara Engelhardt, Sivan Sabato, and Jonathan Scarlett (eds.), \emph{Proceedings of the 40th International Conference on Machine Learning}, volume 202 of \emph{Proceedings of Machine Learning Research}, pp.\  3676--3713. PMLR, 23--29 Jul 2023.
\newblock URL \url{https://proceedings.mlr.press/v202/carta23a.html}.

\bibitem[Chang et~al.(2024)Chang, Wang, Wang, Wu, Yang, Zhu, Chen, Yi, Wang, Wang, et~al.]{chang2024survey}
Yupeng Chang, Xu~Wang, Jindong Wang, Yuan Wu, Linyi Yang, Kaijie Zhu, Hao Chen, Xiaoyuan Yi, Cunxiang Wang, Yidong Wang, et~al.
\newblock A survey on evaluation of large language models.
\newblock \emph{ACM Transactions on Intelligent Systems and Technology}, 15\penalty0 (3):\penalty0 1--45, 2024.

\bibitem[Cooper(2014)]{cooper2014framework}
Seth Cooper.
\newblock \emph{A framework for scientific discovery through video games}.
\newblock Morgan \& Claypool, 2014.

\bibitem[C\^ot\'e et~al.(2018)C\^ot\'e, K\'ad\'ar, Yuan, Kybartas, Barnes, Fine, Moore, Tao, Hausknecht, Asri, Adada, Tay, and Trischler]{cote18textworld}
Marc-Alexandre C\^ot\'e, \'Akos K\'ad\'ar, Xingdi Yuan, Ben Kybartas, Tavian Barnes, Emery Fine, James Moore, Ruo~Yu Tao, Matthew Hausknecht, Layla~El Asri, Mahmoud Adada, Wendy Tay, and Adam Trischler.
\newblock Textworld: A learning environment for text-based games.
\newblock \emph{CoRR}, abs/1806.11532, 2018.

\bibitem[{Covariant AI}(2024)]{covariant_2024_rfm}
{Covariant AI}.
\newblock Rfm-1: A world model that understands physics, 2024.
\newblock URL \url{https://covariant.ai/insights/rfm-1-a-world-model-that-understands-physics/}.

\bibitem[de~Wynter(2024)]{de2024will}
Adrian de~Wynter.
\newblock Will gpt-4 run doom?
\newblock \emph{arXiv preprint arXiv:2403.05468}, 2024.

\bibitem[Dong et~al.(2022)Dong, Li, Dai, Zheng, Wu, Chang, Sun, Xu, and Sui]{dong2022survey}
Qingxiu Dong, Lei Li, Damai Dai, Ce~Zheng, Zhiyong Wu, Baobao Chang, Xu~Sun, Jingjing Xu, and Zhifang Sui.
\newblock A survey on in-context learning.
\newblock \emph{arXiv preprint arXiv:2301.00234}, 2022.

\bibitem[Ecoffet et~al.(2021{\natexlab{a}})Ecoffet, Huizinga, Lehman, Stanley, and Clune]{first_return}
Adrien Ecoffet, Joost Huizinga, Joel Lehman, Kenneth Stanley, and Jeff Clune.
\newblock First return, then explore.
\newblock \emph{Nature}, 590:\penalty0 580--586, 02 2021{\natexlab{a}}.
\newblock \doi{10.1038/s41586-020-03157-9}.

\bibitem[Ecoffet et~al.(2021{\natexlab{b}})Ecoffet, Huizinga, Lehman, Stanley, and Clune]{ecoffet2021goexplore}
Adrien Ecoffet, Joost Huizinga, Joel Lehman, Kenneth~O. Stanley, and Jeff Clune.
\newblock Go-explore: a new approach for hard-exploration problems, 2021{\natexlab{b}}.

\bibitem[Gallouédec \& Dellandréa(2023)Gallouédec and Dellandréa]{gallouédec2023cellfree}
Quentin Gallouédec and Emmanuel Dellandréa.
\newblock Cell-free latent go-explore, 2023.

\bibitem[Gandhi et~al.(2024)Gandhi, Lee, Grand, Liu, Cheng, Sharma, and Goodman]{gandhi2024stream}
Kanishk Gandhi, Denise Lee, Gabriel Grand, Muxin Liu, Winson Cheng, Archit Sharma, and Noah~D. Goodman.
\newblock Stream of search (sos): Learning to search in language, 2024.

\bibitem[Hu \& Clune(2024)Hu and Clune]{hu2023ThoughtCloning}
Shengran Hu and Jeff Clune.
\newblock {Thought Cloning}: Learning to think while acting by imitating human thinking.
\newblock \emph{Advances in Neural Information Processing Systems}, 36, 2024.

\bibitem[Huang et~al.(2022)Huang, Abbeel, Pathak, and Mordatch]{huang2022language}
Wenlong Huang, Pieter Abbeel, Deepak Pathak, and Igor Mordatch.
\newblock Language models as zero-shot planners: Extracting actionable knowledge for embodied agents, 2022.

\bibitem[Hussein et~al.(2017)Hussein, Gaber, Elyan, and Jayne]{hussein2017imitation}
Ahmed Hussein, Mohamed~Medhat Gaber, Eyad Elyan, and Chrisina Jayne.
\newblock Imitation learning: A survey of learning methods.
\newblock \emph{ACM Comput. Surv.}, 50\penalty0 (2), apr 2017.
\newblock ISSN 0360-0300.
\newblock \doi{10.1145/3054912}.
\newblock URL \url{https://doi.org/10.1145/3054912}.

\bibitem[Jiang et~al.(2023)Jiang, Rockt{\"a}schel, and Grefenstette]{jiang2023general}
Minqi Jiang, Tim Rockt{\"a}schel, and Edward Grefenstette.
\newblock General intelligence requires rethinking exploration.
\newblock \emph{Royal Society Open Science}, 10\penalty0 (6):\penalty0 230539, 2023.

\bibitem[Klissarov et~al.(2023)Klissarov, D'Oro, Sodhani, Raileanu, Bacon, Vincent, Zhang, and Henaff]{klissarov2023motif}
Martin Klissarov, Pierluca D'Oro, Shagun Sodhani, Roberta Raileanu, Pierre-Luc Bacon, Pascal Vincent, Amy Zhang, and Mikael Henaff.
\newblock Motif: Intrinsic motivation from artificial intelligence feedback, 2023.

\bibitem[Krishnamurthy et~al.(2024)Krishnamurthy, Harris, Foster, Zhang, and Slivkins]{krishnamurthy2024large}
Akshay Krishnamurthy, Keegan Harris, Dylan~J. Foster, Cyril Zhang, and Aleksandrs Slivkins.
\newblock Can large language models explore in-context?, 2024.

\bibitem[K{\"u}ttler et~al.(2020)K{\"u}ttler, Nardelli, Miller, Raileanu, Selvatici, Grefenstette, and Rockt{\"a}schel]{kuttler2020nethack}
Heinrich K{\"u}ttler, Nantas Nardelli, Alexander Miller, Roberta Raileanu, Marco Selvatici, Edward Grefenstette, and Tim Rockt{\"a}schel.
\newblock The nethack learning environment.
\newblock \emph{Advances in Neural Information Processing Systems}, 33:\penalty0 7671--7684, 2020.

\bibitem[Ladosz et~al.(2022)Ladosz, Weng, Kim, and Oh]{ladosz2022exploration}
Pawel Ladosz, Lilian Weng, Minwoo Kim, and Hyondong Oh.
\newblock Exploration in deep reinforcement learning: A survey.
\newblock \emph{Information Fusion}, 85:\penalty0 1--22, 2022.

\bibitem[Laurent et~al.(2024)Laurent, Janizek, Ruzo, Hinks, Hammerling, Narayanan, Ponnapati, White, and Rodriques]{laurent2024labbenchmeasuringcapabilitieslanguage}
Jon~M. Laurent, Joseph~D. Janizek, Michael Ruzo, Michaela~M. Hinks, Michael~J. Hammerling, Siddharth Narayanan, Manvitha Ponnapati, Andrew~D. White, and Samuel~G. Rodriques.
\newblock Lab-bench: Measuring capabilities of language models for biology research, 2024.
\newblock URL \url{https://arxiv.org/abs/2407.10362}.

\bibitem[Lee et~al.(2024)Lee, Phatale, Mansoor, Lu, Mesnard, Ferret, Bishop, Hall, Carbune, and Rastogi]{lee2024rlaif}
Harrison Lee, Samrat Phatale, Hassan Mansoor, Kellie~Ren Lu, Thomas Mesnard, Johan Ferret, Colton Bishop, Ethan Hall, Victor Carbune, and Abhinav Rastogi.
\newblock {RLAIF}: Scaling reinforcement learning from human feedback with {AI} feedback, 2024.
\newblock URL \url{https://openreview.net/forum?id=AAxIs3D2ZZ}.

\bibitem[Lehnert et~al.(2024)Lehnert, Sukhbaatar, Su, Zheng, Mcvay, Rabbat, and Tian]{lehnert2024a}
Lucas Lehnert, Sainbayar Sukhbaatar, DiJia Su, Qinqing Zheng, Paul Mcvay, Michael Rabbat, and Yuandong Tian.
\newblock Beyond a*: Better planning with transformers via search dynamics bootstrapping, 2024.

\bibitem[Lewis et~al.(2020)Lewis, Perez, Piktus, Petroni, Karpukhin, Goyal, K\"{u}ttler, Lewis, Yih, Rockt\"{a}schel, Riedel, and Kiela]{NEURIPS2020_6b493230}
Patrick Lewis, Ethan Perez, Aleksandra Piktus, Fabio Petroni, Vladimir Karpukhin, Naman Goyal, Heinrich K\"{u}ttler, Mike Lewis, Wen-tau Yih, Tim Rockt\"{a}schel, Sebastian Riedel, and Douwe Kiela.
\newblock Retrieval-augmented generation for knowledge-intensive nlp tasks.
\newblock In H.~Larochelle, M.~Ranzato, R.~Hadsell, M.F. Balcan, and H.~Lin (eds.), \emph{Advances in Neural Information Processing Systems}, volume~33, pp.\  9459--9474. Curran Associates, Inc., 2020.
\newblock URL \url{https://proceedings.neurips.cc/paper_files/paper/2020/file/6b493230205f780e1bc26945df7481e5-Paper.pdf}.

\bibitem[Liu et~al.(2023)Liu, Yu, Zhang, Xu, Lei, Lai, Gu, Ding, Men, Yang, Zhang, Deng, Zeng, Du, Zhang, Shen, Zhang, Su, Sun, Huang, Dong, and Tang]{liu2023agentbench}
Xiao Liu, Hao Yu, Hanchen Zhang, Yifan Xu, Xuanyu Lei, Hanyu Lai, Yu~Gu, Hangliang Ding, Kaiwen Men, Kejuan Yang, Shudan Zhang, Xiang Deng, Aohan Zeng, Zhengxiao Du, Chenhui Zhang, Sheng Shen, Tianjun Zhang, Yu~Su, Huan Sun, Minlie Huang, Yuxiao Dong, and Jie Tang.
\newblock Agentbench: Evaluating llms as agents, 2023.

\bibitem[{Llama Team}(2024)]{llama3}
{Llama Team}.
\newblock The llama 3 herd of models, 2024.
\newblock URL \url{https://arxiv.org/abs/2407.21783}.

\bibitem[Lu et~al.(2024)Lu, Georgescu, and Verwey]{lu_2024_3d}
Cong Lu, Raluca Georgescu, and Johan Verwey.
\newblock Go-explore complex 3-d game environments for automated reachability testing.
\newblock \emph{IEEE Transactions on Games}, 16\penalty0 (1):\penalty0 235--240, 2024.
\newblock \doi{10.1109/TG.2022.3228401}.

\bibitem[Madotto et~al.(2020)Madotto, Namazifar, Huizinga, Molino, Ecoffet, Zheng, Papangelis, Yu, Khatri, and Tur]{madotto2020exploration}
Andrea Madotto, Mahdi Namazifar, Joost Huizinga, Piero Molino, Adrien Ecoffet, Huaixiu Zheng, Alexandros Papangelis, Dian Yu, Chandra Khatri, and Gokhan Tur.
\newblock Exploration based language learning for text-based games.
\newblock \emph{arXiv preprint arXiv:2001.08868}, 2020.

\bibitem[Olsson et~al.(2022)Olsson, Elhage, Nanda, Joseph, DasSarma, Henighan, Mann, Askell, Bai, Chen, et~al.]{olsson2022context}
Catherine Olsson, Nelson Elhage, Neel Nanda, Nicholas Joseph, Nova DasSarma, Tom Henighan, Ben Mann, Amanda Askell, Yuntao Bai, Anna Chen, et~al.
\newblock In-context learning and induction heads.
\newblock \emph{arXiv preprint arXiv:2209.11895}, 2022.

\bibitem[OpenAI(2024)]{openai2024gpt4}
OpenAI.
\newblock Gpt-4 technical report, 2024.

\bibitem[Ouyang et~al.(2022)Ouyang, Wu, Jiang, Almeida, Wainwright, Mishkin, Zhang, Agarwal, Slama, Ray, Schulman, Hilton, Kelton, Miller, Simens, Askell, Welinder, Christiano, Leike, and Lowe]{ouyang2022traininglanguagemodelsfollow}
Long Ouyang, Jeff Wu, Xu~Jiang, Diogo Almeida, Carroll~L. Wainwright, Pamela Mishkin, Chong Zhang, Sandhini Agarwal, Katarina Slama, Alex Ray, John Schulman, Jacob Hilton, Fraser Kelton, Luke Miller, Maddie Simens, Amanda Askell, Peter Welinder, Paul Christiano, Jan Leike, and Ryan Lowe.
\newblock Training language models to follow instructions with human feedback, 2022.
\newblock URL \url{https://arxiv.org/abs/2203.02155}.

\bibitem[Raad et~al.(2024)Raad, Ahuja, Barros, Besse, Bolt, Bolton, Brownfield, Buttimore, Cant, Chakera, et~al.]{raad2024scaling}
Maria~Abi Raad, Arun Ahuja, Catarina Barros, Frederic Besse, Andrew Bolt, Adrian Bolton, Bethanie Brownfield, Gavin Buttimore, Max Cant, Sarah Chakera, et~al.
\newblock Scaling instructable agents across many simulated worlds.
\newblock \emph{arXiv preprint arXiv:2404.10179}, 2024.

\bibitem[Shinn et~al.(2023)Shinn, Cassano, Berman, Gopinath, Narasimhan, and Yao]{shinn2023reflexion}
Noah Shinn, Federico Cassano, Edward Berman, Ashwin Gopinath, Karthik Narasimhan, and Shunyu Yao.
\newblock Reflexion: Language agents with verbal reinforcement learning, 2023.

\bibitem[Sutton \& Barto(2018)Sutton and Barto]{Sutton1998}
Richard~S. Sutton and Andrew~G. Barto.
\newblock \emph{Reinforcement Learning: An Introduction}.
\newblock The MIT Press, second edition, 2018.
\newblock URL \url{http://incompleteideas.net/book/the-book-2nd.html}.

\bibitem[Talmor et~al.(2019)Talmor, Herzig, Lourie, and Berant]{talmor2019commonsense}
Alon Talmor, Jonathan Herzig, Nicholas Lourie, and Jonathan Berant.
\newblock {C}ommonsense{QA}: A question answering challenge targeting commonsense knowledge.
\newblock In Jill Burstein, Christy Doran, and Thamar Solorio (eds.), \emph{Proceedings of the 2019 Conference of the North {A}merican Chapter of the Association for Computational Linguistics: Human Language Technologies, Volume 1 (Long and Short Papers)}, pp.\  4149--4158, Minneapolis, Minnesota, June 2019. Association for Computational Linguistics.
\newblock \doi{10.18653/v1/N19-1421}.
\newblock URL \url{https://aclanthology.org/N19-1421}.

\bibitem[Team(2024)]{geminiteam2024gemini}
Gemini Team.
\newblock Gemini: A family of highly capable multimodal models, 2024.

\bibitem[Torne~Villasevil et~al.(2023)Torne~Villasevil, Balsells I~Pamies, Wang, Desai, Chen, Agrawal, and Gupta]{NEURIPS2023_c7c7cf10}
Marcel Torne~Villasevil, Max Balsells I~Pamies, Zihan Wang, Samedh Desai, Tao Chen, Pulkit Agrawal, and Abhishek Gupta.
\newblock Breadcrumbs to the goal: Goal-conditioned exploration from human-in-the-loop feedback.
\newblock In A.~Oh, T.~Neumann, A.~Globerson, K.~Saenko, M.~Hardt, and S.~Levine (eds.), \emph{Advances in Neural Information Processing Systems}, volume~36, pp.\  63222--63258. Curran Associates, Inc., 2023.
\newblock URL \url{https://proceedings.neurips.cc/paper_files/paper/2023/file/c7c7cf10082e454b9662a686ce6f1b6f-Paper-Conference.pdf}.

\bibitem[Touvron et~al.(2023)Touvron, Martin, Stone, Albert, Almahairi, Babaei, Bashlykov, Batra, Bhargava, Bhosale, et~al.]{touvron2023llama}
Hugo Touvron, Louis Martin, Kevin Stone, Peter Albert, Amjad Almahairi, Yasmine Babaei, Nikolay Bashlykov, Soumya Batra, Prajjwal Bhargava, Shruti Bhosale, et~al.
\newblock Llama 2: Open foundation and fine-tuned chat models.
\newblock \emph{arXiv preprint arXiv:2307.09288}, 2023.

\bibitem[Trinh et~al.(2024)Trinh, Wu, Le, He, and Luong]{trinh2024solving}
Trieu~H Trinh, Yuhuai Wu, Quoc~V Le, He~He, and Thang Luong.
\newblock Solving olympiad geometry without human demonstrations.
\newblock \emph{Nature}, 625\penalty0 (7995):\penalty0 476--482, 2024.

\bibitem[Wang et~al.(2023)Wang, Xie, Jiang, Mandlekar, Xiao, Zhu, Fan, and Anandkumar]{wang2023voyager}
Guanzhi Wang, Yuqi Xie, Yunfan Jiang, Ajay Mandlekar, Chaowei Xiao, Yuke Zhu, Linxi Fan, and Anima Anandkumar.
\newblock Voyager: An open-ended embodied agent with large language models, 2023.

\bibitem[Wang et~al.(2024)Wang, Ma, Feng, Zhang, Yang, Zhang, Chen, Tang, Chen, Lin, et~al.]{wang2024survey}
Lei Wang, Chen Ma, Xueyang Feng, Zeyu Zhang, Hao Yang, Jingsen Zhang, Zhiyuan Chen, Jiakai Tang, Xu~Chen, Yankai Lin, et~al.
\newblock A survey on large language model based autonomous agents.
\newblock \emph{Frontiers of Computer Science}, 18\penalty0 (6):\penalty0 1--26, 2024.

\bibitem[Wei et~al.(2022)Wei, Wang, Schuurmans, Bosma, Xia, Chi, Le, Zhou, et~al.]{wei2022chain}
Jason Wei, Xuezhi Wang, Dale Schuurmans, Maarten Bosma, Fei Xia, Ed~Chi, Quoc~V Le, Denny Zhou, et~al.
\newblock Chain-of-thought prompting elicits reasoning in large language models.
\newblock \emph{Advances in neural information processing systems}, 35:\penalty0 24824--24837, 2022.

\bibitem[Yao et~al.(2023{\natexlab{a}})Yao, Yu, Zhao, Shafran, Griffiths, Cao, and Narasimhan]{yao2023tree}
Shunyu Yao, Dian Yu, Jeffrey Zhao, Izhak Shafran, Thomas~L. Griffiths, Yuan Cao, and Karthik Narasimhan.
\newblock {Tree of Thoughts}: Deliberate problem solving with large language models, 2023{\natexlab{a}}.

\bibitem[Yao et~al.(2023{\natexlab{b}})Yao, Zhao, Yu, Du, Shafran, Narasimhan, and Cao]{yao2023react}
Shunyu Yao, Jeffrey Zhao, Dian Yu, Nan Du, Izhak Shafran, Karthik~R Narasimhan, and Yuan Cao.
\newblock React: Synergizing reasoning and acting in language models.
\newblock In \emph{The Eleventh International Conference on Learning Representations}, 2023{\natexlab{b}}.
\newblock URL \url{https://openreview.net/forum?id=WE_vluYUL-X}.

\bibitem[Zhang et~al.(2024)Zhang, Lehman, Stanley, and Clune]{zhang2024omni}
Jenny Zhang, Joel Lehman, Kenneth Stanley, and Jeff Clune.
\newblock {OMNI}: Open-endedness via models of human notions of interestingness.
\newblock In \emph{The Twelfth International Conference on Learning Representations}, 2024.
\newblock URL \url{https://openreview.net/forum?id=AgM3MzT99c}.

\bibitem[Zhao et~al.(2021)Zhao, Wallace, Feng, Klein, and Singh]{zhao2021calibrate}
Zihao Zhao, Eric Wallace, Shi Feng, Dan Klein, and Sameer Singh.
\newblock Calibrate before use: Improving few-shot performance of language models.
\newblock In \emph{International conference on machine learning}, pp.\  12697--12706. PMLR, 2021.

\bibitem[Zheng et~al.(2023{\natexlab{a}})Zheng, Chiang, Sheng, Zhuang, Wu, Zhuang, Lin, Li, Li, Xing, Zhang, Gonzalez, and Stoica]{NEURIPS2023_91f18a12}
Lianmin Zheng, Wei-Lin Chiang, Ying Sheng, Siyuan Zhuang, Zhanghao Wu, Yonghao Zhuang, Zi~Lin, Zhuohan Li, Dacheng Li, Eric Xing, Hao Zhang, Joseph~E Gonzalez, and Ion Stoica.
\newblock Judging llm-as-a-judge with mt-bench and chatbot arena.
\newblock In A.~Oh, T.~Naumann, A.~Globerson, K.~Saenko, M.~Hardt, and S.~Levine (eds.), \emph{Advances in Neural Information Processing Systems}, volume~36, pp.\  46595--46623. Curran Associates, Inc., 2023{\natexlab{a}}.
\newblock URL \url{https://proceedings.neurips.cc/paper_files/paper/2023/file/91f18a1287b398d378ef22505bf41832-Paper-Datasets_and_Benchmarks.pdf}.

\bibitem[Zheng et~al.(2023{\natexlab{b}})Zheng, Chiang, Sheng, Zhuang, Wu, Zhuang, Lin, Li, Li, Xing, Zhang, Gonzalez, and Stoica]{zheng2023judging}
Lianmin Zheng, Wei-Lin Chiang, Ying Sheng, Siyuan Zhuang, Zhanghao Wu, Yonghao Zhuang, Zi~Lin, Zhuohan Li, Dacheng Li, Eric~P. Xing, Hao Zhang, Joseph~E. Gonzalez, and Ion Stoica.
\newblock Judging llm-as-a-judge with mt-bench and chatbot arena, 2023{\natexlab{b}}.

\bibitem[Zhu et~al.(2023{\natexlab{a}})Zhu, Chen, Shen, Li, and Elhoseiny]{zhu2023minigpt}
Deyao Zhu, Jun Chen, Xiaoqian Shen, Xiang Li, and Mohamed Elhoseiny.
\newblock Minigpt-4: Enhancing vision-language understanding with advanced large language models.
\newblock \emph{arXiv preprint arXiv:2304.10592}, 2023{\natexlab{a}}.

\bibitem[Zhu et~al.(2023{\natexlab{b}})Zhu, Chen, Tian, Tao, Su, Yang, Huang, Li, Lu, Wang, Qiao, Zhang, and Dai]{zhu2023ghost}
Xizhou Zhu, Yuntao Chen, Hao Tian, Chenxin Tao, Weijie Su, Chenyu Yang, Gao Huang, Bin Li, Lewei Lu, Xiaogang Wang, Yu~Qiao, Zhaoxiang Zhang, and Jifeng Dai.
\newblock Ghost in the minecraft: Generally capable agents for open-world environments via large language models with text-based knowledge and memory.
\newblock \emph{arXiv preprint arXiv:2305.17144}, 2023{\natexlab{b}}.

\bibitem[Zoubir \& Iskandler(2007)Zoubir and Iskandler]{zoubir2007bootstrap}
Abdefihak~M Zoubir and D~Robert Iskandler.
\newblock Bootstrap methods and applications.
\newblock \emph{IEEE Signal Processing Magazine}, 24\penalty0 (4):\penalty0 10--19, 2007.

\end{thebibliography}
\bibliographystyle{include/iclr2025_conference}

\clearpage
\appendix

\section*{\LARGE Supplementary Material}

\vspace*{20pt}
\section*{Table of Contents}
\vspace*{-5pt}
\startcontents[sections]
\printcontents[sections]{l}{1}{\setcounter{tocdepth}{2}}

\clearpage

\section{Algorithm Pseudocode}
\label{appsec:pseudocode}

We provide full pseudocode for \ouralgolong in \Cref{alg:main_alg}.
This complements the discussion in \Cref{sec:algo}.

\begin{algorithm}[h!]
\begin{footnotesize}
\caption{\ouralgolong}
\label{alg:main_alg}
\begin{algorithmic}[1]
\State \textbf{Hyperparameters:} no.\ state expansions $N_{\text{state}}$, no.\ exploratory actions $N_{\text{action}}$, foundation model $\mathcal{M}$
\State \textbf{Initialize:} archive of states $\mathcal{S}_{\textrm{archive}}=\emptyset$, state-conditional action history $\mathcal{A}(\cdot) =\emptyset$
\State $\mathcal{S}_{\textrm{archive}} \gets \mathcal{S}_{\textrm{archive}} \cup \{s_0\}$ \Comment{Add initial state to archive}
\For{$i=1, \dots, N_{\textrm{state}}$}
    \State Query $\mathcal{M}$ for the next state $s_{i,1}$ from $\mathcal{S}_{\textrm{archive}}$ \Comment{See \Cref{subsec:ige_select_state}}
    \For{$j=1, \dots, N_{\textrm{action}}$}
        \State Query $\mathcal{M}$ for the next action $a_{i,j}$ from $s_{i,j}$ conditional on $\mathcal{A}(s_{i,j})$ \Comment{See \Cref{subsec:ige_select_action}}
        \State $s_{i,j+1} \sim P(s_{i,j}, a_{i,j})$, $\mathcal{A}(s_{i,j}) \gets \mathcal{A}(s_{i,j}) \cup \{a_{i,j}\}$ \Comment{Take action and update history}
        \If{$\mathcal{M}$ determines that $s_{i,j+1}$ is interesting w.r.t.\ $\mathcal{S}_{\textrm{archive}}$} \Comment{See \Cref{subsec:ige_filter_state}}
            \State $\mathcal{S}_{\textrm{archive}} \gets \mathcal{S}_{\textrm{archive}} \cup \{s_{i,j+1}\}$
        \EndIf
    \EndFor
\EndFor
\State Return best discovered trajectory
\end{algorithmic}
\end{footnotesize}
\end{algorithm}

\section{Further Details on Environments}
\label{appsec:envs}

We provide further details for each of the environments used in the empirical evaluation in \Cref{sec:eval}, and the environment-specific descriptions appended to the system prompts.
Each environment description may include high-level information about the task and a description of the action space.

\subsection{Game of 24}
\label{appsubsec:game24}

We use the environment and set of evaluation tasks from \url{https://github.com/princeton-nlp/tree-of-thought-llm} which is released under the MIT License.
We include the environment-specific prompt that is appended to the system prompt in \Cref{sec:algo} below.
The system prompt contains examples of correct reasoning paths on different problems (few-shot prompting).

\begin{mybox}{Environment Description.}
\small
You are given 4 numbers and must use basic arithmetic operations (+ - * /) to obtain 24.
At each step, you are only allowed to choose two of the remaining numbers to obtain a new number.
A correct answer is one that uses each input exactly once and no other numbers.
Reaching 24 before the last step does not count as a correct answer.
Follow the convention that division is integer division, and never by zero.
Some examples of correct reasoning traces are as follows:\\
Initial state: (4 4 6 8)\\
Steps:\\
4 + 8 = 12. Next: (4 6 12)\\
6 - 4 = 2. Next: (2 12)\\
2 * 12 = 24. Next: (24)\\
Answer: (6 - 4) * (4 + 8) = 24\\
Initial state: (2 9 10 12)\\
Steps:\\
12 * 2 = 24. Next: (9 10 24)\\
10 - 9 = 1. Next: (1 24)\\
24 * 1 = 24. Next: (24)\\
Answer: (12 * 2) * (10 - 9) = 24\\
Initial state: (4 9 10 13)\\
Steps:\\
13 - 10 = 3. Next: (3 4 9)\\
9 - 3 = 6. Next: (4 6)\\
4 * 6 = 24. Next: (24)\\
Answer: 4 * (9 - (13 - 10)) = 24\\
\end{mybox}

The action space at each step is all the valid arithmetic operations, presented in an analogous way as the `propose' step in \citet{yao2023tree}.

\subsection{BabyAI}
\label{appsubsec:babyai}
\begin{figure}[h!]
\vspace{-2mm}
\centering
\includegraphics[width=0.99\textwidth, trim={130 70 100 50}, clip]{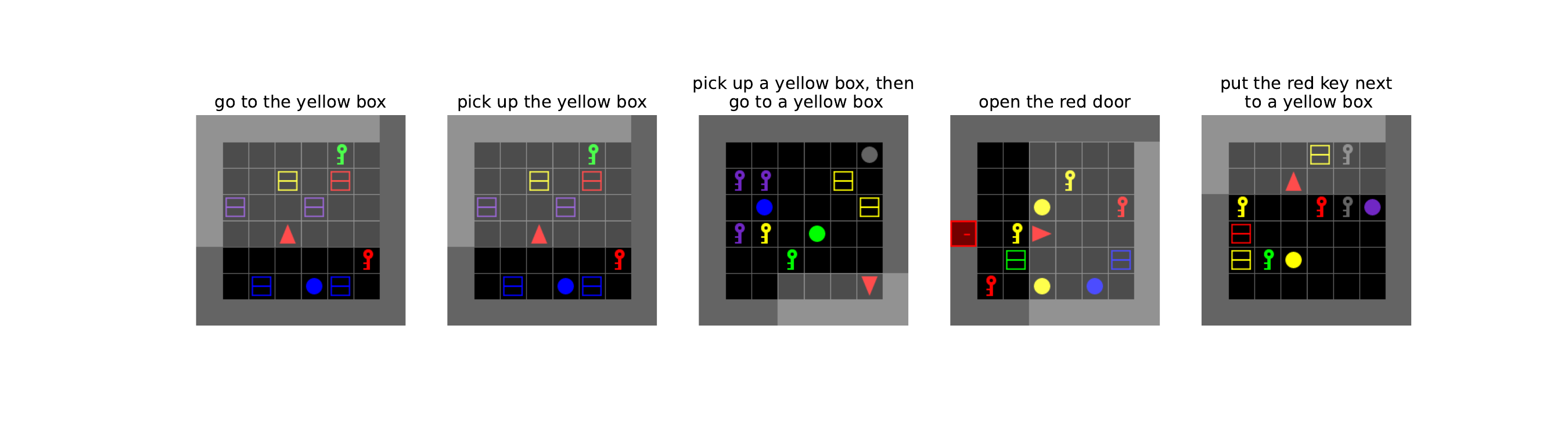}
\vspace{-2mm}
\caption{\small{
We visualize the 5 types of tasks that BabyAI consists of for our evaluation in \Cref{subsec:eval_babyai}. \ouralgo receives only partial text-based observations corresponding to the view in the figure.
}}
\label{fig:babyai_env_spread}
\end{figure}

The BabyAI~\citep{pmlr-v202-carta23a} environment comes with five task types, which we list here and visualize in order in \Cref{fig:babyai_env_spread}:
\begin{itemize}
    \item
    Go to $<$object$>$, a simple navigation task that requires reasoning abilities to choose the right plan given the object's position;
    \item 
    Pick up $<$object$>$, a reasoning task that combines navigation tasks;
    \item
    Pick up $<$object A$>$ then go to $<$object B$>$ and Go to $<$object B$>$ after pickup $<$object A$>$, both serving to test reasoning abilities on temporal sequences;
    \item 
    Unlock $<$door$>$, a task that includes inferring that a key is needed to unlock the door, finding the right key (i.e., the one colored as the door), and eventually using the toggle action with the key on the door;
    \item 
    Put $<$object A$>$ next to $<$object B$>$, which requires first reaching $<$object A$>$, picking it up, reaching $<$object B$>$ and finally dropping $<$object A$>$ next to $<$object B$>$.
\end{itemize}

We use the codebase from \url{https://github.com/flowersteam/Grounding_LLMs_with_online_RL} which is released under the MIT License.
The action space is discrete and composed of 6 possible actions: turn left, turn right, go forward, pick up, drop, and toggle.
The `go to' and `pick up' tasks have a shorter environment horizon of $H=64$, whereas the rest have a horizon of $H=128$.
We include the environment-specific prompt that is appended to the system prompt in \Cref{sec:algo} below.

\begin{mybox}{Environment Description (Text).}
\small
You are an agent in an 8x8 partially-observable 2D text-based environment.
You see the 6x6 grid in front of you, and can face north, south, east, or west.
The possible actions are turn left, turn right, go forward, pick up, drop, and toggle.
At each turn, you will receive a history of the last observations and actions.
Your aim is to complete the task described in the goal.
Each tile in the grid can only contain at most one object.
Objects cannot be crossed, and may need to be bypassed or moved.
You can only move onto an empty tile or on a tile containing an open door.
You can only hold one object at a time, using pick up when they are one step in front.
Objects are dropped one tile in front and cannot be dropped when there is another object in front.
Doors are unlocked with keys of the same color using the toggle action.
Actions are deterministic, do not repeat actions if they have no effect.
You have H steps to complete the task.
\end{mybox}

\begin{mybox}{Environment Description (Visual).}
\small
You are an agent in an 8x8 partially-observable 2D image-based environment.
You see the 6x6 grid in front of you, and can face north, south, east, or west.
The possible actions are turn left, turn right, go forward, pick up, drop, and toggle.
At each turn, you will receive a history of the last observations and actions.
The observations will be presented as a sequence of images.
Your aim is to complete the task described in the goal.
Each tile in the grid can only contain at most one object.
Objects cannot be crossed, and may need to be bypassed or moved.
You can only move onto an empty tile or on a tile containing an open door.
You can only hold one object at a time, using pick up when they are one step in front.
Objects are dropped one tile in front and cannot be dropped when there is another object in front.
Doors are unlocked with keys of the same color using the toggle action.
Actions are deterministic, do not repeat actions if they have no effect.
You have H steps to complete the task.
\end{mybox}

\subsection{TextWorld}
\label{appsubsec:textworld}
We evaluate \ouralgo on `Treasure Hunter', `The Cooking Game', and `Coin Collector' from the TextWorld~\citep{cote18textworld} domain.
We use the environment code from \url{https://github.com/microsoft/TextWorld} which is released under the MIT License.

\subsubsection{Treasure Hunter}
For Treasure Hunter, we set the `level' option to the maximum value of 30, resulting in a maze with 20 rooms.
Locked doors and containers are added, which may need to be unlocked and opened to find the target object.
To further increase the difficulty, we remove the solution description from the original game and filter out tasks that can be completed with 20 steps in the optimal solution.
We include the environment-specific prompt that is appended to the system prompt in \Cref{sec:algo} below.

\begin{mybox}{Environment Description for Treasure Hunter.}
\small
You are an agent playing TextWorld, a text-based adventure game where you are in a randomly generated maze and must find a specific object. You need to explore different rooms to find the target object.

Here are the available commands:
  look:                describe the current room.
  goal:                print the goal of this game
  inventory:           print the player's inventory
  go $<$dir$>$:            move the player north, east, south, or west. You can only go in the direction indicated with an exit or a door.
  open ...:            open a door or a container. You need to open a closed door before you want to go through it.
  drop ...:            drop an object on the floor
  take ...:            take an object that is visible. Make sure the object is visible to take.
  put ... on ...:      place an object on a supporter
  take ... from ...:   take an object from a container or a supporter
  insert ... into ...: place an object into a container
  unlock ... with ...: unlock a door or a container with a key. You need to unlock a locked door with a matched key in your inventory before you want to open it.

- The target object might be located in a closed or locked container. 
- The adjective is useful for determining whether the key is matched with the lock (e.g., non-euclidean keycard is matched with non-euclidean safe). Make sure it is matched to unlock!
- The key required to unlock the door may be in another room or locked inside a container.
- Take the key whenever you can.
- After unlocking a locked door or container, it will remain closed. You will then need to open it.

You have 40 steps to complete the task. Restarting is forbidden.
\end{mybox}

\subsubsection{The Cooking Game}

In The Cooking Game, we set the number of ingredients to a maximum of 5 and the number of rooms to 13.
We enable all challenging additional options: doors need to be opened, food must be processed (e.g., cut, diced, chopped with a knife), and cooked (e.g., grilled with a BBQ, fried on a stove, roasted in an oven).
We include the environment-specific prompt that is appended to the system prompt in \Cref{sec:algo} below.

\begin{mybox}{Environment Description for The Cooking Game.}
\small
You are an agent playing TextWorld, a text-based adventure game where you navigate through different rooms, interact with objects, and solve puzzles. 
Your goal is to first find the recipe, find and prepare food according to the recipe, and finally prepare and eat the meal.

Here are the available commands:
  look:                describe the current room
  goal:                print the goal of this game
  inventory:           print player's inventory
  go $<$dir$>$:            move the player north, east, south or west. You can only go to directions indicated with an exit or a door.
  examine ...:         examine something more closely
  eat ...:             eat edible food
  open ...:            open a door or a container. You need to open a closed door before you can go through it.
  drop ...:            drop an object onto the floor
  take ...:            take an object that is visible
  put ... on ...:      place an object on a supporter
  take ... from ...:   take an object from a container or a supporter
  insert ... into ...: place an object into a container
  lock ... with ...:   lock a door or a container with a key
  unlock ... with ...: unlock a door or a container with a key
  cook ... with ...:   cook cookable food with something providing heat
  slice ... with ...:  slice cuttable food with something sharp
  chop ... with ...:   chop cuttable food with something sharp
  dice ... with ...:   dice cuttable food with something sharp
  prepare meal:        combine ingredients from inventory into a meal. You can only prepare meals in the Kitchen.
  
- You can examine the cookbook to see the recipe when it is visible.
- The BBQ is for grilling things, the stove is for frying things, the oven is for roasting things. Cooking ingredients in the wrong way will lead to a failure of the game.
- Once you have got processed ingredients and the appropriate cooking tool ready, cook all of them according to the recipe.
- There are two conditions to correctly cook something (grill/fry/roast): a) the ingredient you want to cook is in your inventory and b) there is a suitable cooking tool in the room, and then use `cook ... with ...' command.
- When you need to chop/slice/dice ingredients, you need to take the knife and the ingredient in your inventory and then `slice/chop/dice ... with knife'
- Make sure to first process the food (chop/slice/dice) before you try to cook them.
- When you have all the ingredients (that got processed or cooked according to the menu), you can `prepare meal' in the kitchen and then `eat meal' to win the game.
- The ingredients should EXACTLY match the color in the recipe, but if the recipe doesn't specify color, any color would be fine. When you `take ... with ...', use the EXACT name you see.
- You don't need to examine the container/supporter (e.g., toolbox) when it says something like "there isn't a thing on it"/"has nothing on it" 

You have 80 steps to complete the task. Restarting is forbidden.
\end{mybox}

We show a successful example trajectory found by \ouralgo below, from our evaluation in \Cref{subsec:eval_textworld}.

\begin{mybox}{Example trajectory found by \ouralgo in The Cooking Game.}
\small
You are hungry! Let's cook a delicious meal. Check the cookbook in the kitchen for the recipe. Once done, enjoy your meal!

-= Backyard =-
You find yourself in a backyard. A normal kind of place. You can see a patio table. But oh no! there's nothing on this piece of garbage. This always happens! You make out a patio chair. But oh no! there's nothing on this piece of trash. You can make out a BBQ, so there's that. There is a closed barn door leading north. There is a closed sliding patio door leading west. There is an exit to the east. Don't worry, there is no door.

$>$ open sliding patio door

You open sliding patio door.

$>$ go west

-= Kitchen =-
You find yourself in a kitchen. An ordinary kind of place. The room is well lit.
You can make out a fridge. You hear a noise behind you and spin around, but you can't see anything other than an oven. Were you looking for a table? Because look over there, it's a table. On the table you see a cookbook. You can't wait to tell the folks at home about this! Look out! It's a- oh, never mind, it's just a counter. Now why would someone leave that there? You see a red apple, a raw purple potato, a red onion, a yellow apple and a knife on the counter. You can see a stove. You shudder, but continue examining the stove. Unfortunately, there isn't a thing on it. Hm. Oh well
There is an open sliding patio door leading east. There is a closed plain door leading north. You don't like doors? Why not try going west, that entranceway is not blocked by one.

$>$ examine cookbook

You open the copy of "Cooking: A Modern Approach (3rd Ed.)" and start reading:

Recipe \#1
---------
Gather all following ingredients and follow the directions to prepare this tasty meal.
Ingredients:
block of cheese
  carrot
  orange bell pepper
  red apple
  tomato
Directions:
chop the block of cheese
  fry the block of cheese
  dice the carrot
  grill the carrot
  chop the orange bell pepper
  fry the orange bell pepper
  chop the red apple
  grill the red apple
  slice the tomato
  grill the tomato
  prepare meal

$>$ take red apple

You take the red apple from the counter.

Your score has just gone up by one point.

$>$ open fridge

You open the fridge, revealing a block of cheese, a carrot, a white onion and a yellow bell pepper.

$>$ take block of cheese

You take the block of cheese from the fridge.

Your score has just gone up by one point.

$>$ take carrot

You take the carrot from the fridge.

Your score has just gone up by one point.

$>$ examine counter

The counter is solid.

On the counter are a raw purple potato, a red onion, a yellow apple and a knife.

$>$ go west

-= Livingroom =-
You arrive in a livingroom...

... (skip 30 steps)

$>$ cook tomato with BBQ

You grilled the tomato.

Your score has just gone up by one point.

$>$ go west

-= Kitchen =-
You find yourself in a kitchen...

$>$ prepare meal

Adding the meal to your inventory.

Your score has just gone up by one point.

$>$ eat meal

You eat the meal. Not bad.

Your score has just gone up by one point.

                                                       *** The End ***

You scored 17 out of a possible 17, in 44 turns.
\end{mybox}

\subsubsection{Coin Collector}
In Coin Collector, we set the number of rooms to 40 and allow distractor rooms to be added along the way.
Similar to Treasure Hunter, we remove the solution description from the original game, and the optimal path from the agent's starting point to the target is set to 20 steps.
We include the environment-specific prompt that is appended to the system prompt in \Cref{sec:algo} below.

\begin{mybox}{Environment Description for Coin Collector.}
\small
You are an agent playing TextWorld, a text-based adventure game where you are in a randomly generated maze and must find the coin. You need to explore different rooms to find the target object.

Here are the available commands:
  goal:                print the goal of this game
  go $<$dir$>$:            move the player north, east, south, or west. You can only go in the direction indicated with something like an exit or a door.
  take coin:            win the game by `take coin' if you see the coin in the room

The only action you can do is `go $<$dir$>$' to explore the maze and `take coin' when you see the coin in the room.

You have 25 steps to complete the task. Restarting is forbidden.
\end{mybox}

\section{Additional Experiments and Analysis}
\label{appsec:additional_experiments}

\subsection{Archive Size Analysis}
\label{appsec:archive_size}

We provide the detailed analysis of the impact of intelligent archive filtering on the size of the archive across different environments.
As shown in \Cref{tab:archive_size_analysis}, intelligent filtering in \ouralgo can drastically reduce the size of the archive and help the algorithm focus on the most interesting states.
Without intelligent filtering in BabyAI-Text and TextWorld, the archive becomes over $5\times$ larger.

\setlength{\tabcolsep}{21pt}
\begin{table}[h!]
\small
\centering
\vspace{-2mm}
\caption{
\small
We show that intelligent filtering in \ouralgo can drastically reduce the size of the archive, and help the algorithm focus on the most interesting states.
We use the same evaluation setting as \Cref{tab:ige_ablations} and show the mean and 95\% bootstrap confidence intervals.
}
\vspace{-2mm}
\label{tab:archive_size_analysis}
\begin{tabular}{lccc}
\toprule
\multicolumn{1}{c}{\multirow{2}{*}{\textbf{Archive Filtering}}} & \multicolumn{3}{c}{\textbf{Number of States}} \\
\multicolumn{1}{c}{}                                   
& \textbf{Game of 24} & \textbf{BabyAI (PN)} & \textbf{TextWorld (CG)} \\ \midrule
No Filter                                    & 18.5 $\pm$ 3.2                   & 203.5 $\pm$ 56.7   & 22.4 $\pm$ 15.3           \\
With Filter                                      & \textbf{15.6 $\pm$ 2.3}                   & \textbf{25.5 $\pm$ 5.2}   & \textbf{\phantom{0}4.4 $\pm$ 2.8\phantom{0}}              \\ \bottomrule
\end{tabular}
\end{table}

These results highlight that, across all evaluated environments, intelligent archive filtering not only enhances performance but also maintains a manageable archive size, which is crucial for scalability.

\subsection{Prompt Robustness Experiments}
\label{appsubsec:prompt_robustness}

We investigated the sensitivity of \ouralgo to prompt variations to assess its robustness to prompt engineering.
Specifically, we conducted experiments where we removed any hints or domain-specific instructions on how to solve the tasks, keeping the prompts purely factually descriptive about the environment.
For example, in:
\begin{itemize}
    \item Game of 24
    \begin{itemize}
        \item We removed all few-shot prompting, so that the agent has no examples. I.e., all text from ``Some examples of correct reasoning traces are as follows:''
    \end{itemize}
    \item BabyAI
    \begin{itemize}
        \item We removed ``do not repeat actions if they have no effect''
    \end{itemize}
    \item TextWorld (Cooking Game)
    \begin{itemize}
        \item We removed ``The ingredients should EXACTLY match the color in the recipe, but if the recipe doesn't specify color, any color would be fine. When you `take ... with ...', use the EXACT name you see.''
        \item We removed ``You don't need to examine the container/supporter (e.g., toolbox) when it says something like ``there isn't a thing on it''/``has nothing on it'' ''
    \end{itemize}
\end{itemize}
We refer to these as domain-general prompts, as they contain only basic information about the environment without any guidance on how to approach the tasks.

The evaluation settings remained identical to those in our main experiments (see \Cref{sec:eval}), ensuring a fair comparison.
The results, shown in \Cref{tab:ige_prompt_variations}, indicate that even with minimal prompts, \ouralgo maintains a significant performance advantage over the baselines across all environments.
For instance, in the Game of 24, \ouralgo achieves a success rate of $96\%$ with domain-general prompts, compared to $100\%$ with the original prompts, and still significantly outperforms the baselines.
Similar trends are observed in the BabyAI ``Put Next To'' task and the TextWorld ``Cooking Game''.
The performance drop of \ouralgo when using domain-general prompts is modest compared to the baselines, demonstrating that \ouralgo is robust to prompt variations and does not rely heavily on domain-specific prompt engineering.

These findings suggest that \ouralgo can be applied effectively without extensive prompt tuning, making it flexible and generalizable to new domains.
The foundation model's internalized notions of interestingness and problem-solving abilities enable \ouralgo to guide exploration even with minimal guidance, highlighting its potential for widespread applicability.

\setlength{\tabcolsep}{8pt}
\begin{table}[h!]
\small
\centering
\caption{
\small
\ouralgo Performance with Domain-General Prompts.
We use the same evaluation setting as \Cref{tab:ige_ablations} and show the mean and 95\% bootstrap confidence interval for the success rate.
}
\vspace{-2mm}
\label{tab:ige_prompt_variations}
\resizebox{\textwidth}{!}{
\begin{tabular}{llcccc}
\toprule
\textbf{Environment}           & \textbf{Model}            & \textbf{Na\"ive LLM} & \textbf{ReAct} & \textbf{Reflexion} & \textbf{\ouralgo}        \\
\midrule
\multirow{2}{*}{Game of 24}            & Original (GPT-4)     & $70 \pm 14$     & $82 \pm 12$     & $83 \pm 11$    & $\mathbf{100 \pm 0}$ \\
                                   & Domain-General Prompt & $35 \pm 13$     & $71 \pm 13$     & $71 \pm 13$    & $\mathbf{96 \pm 4}$ \\
\midrule
\multirow{2}{*}{BabyAI ``Put Next To''}  & Original (GPT-4)     & $12 \pm 12$     & $48 \pm 12$     & N/A           & $\mathbf{84 \pm 8}$ \\
                                   & Domain-General Prompt & $8 \pm 12$      & $24 \pm 12$     & N/A           & $\mathbf{68 \pm 12}$ \\
\midrule
\multirow{2}{*}{TextWorld ``Cooking Game''} & Original (GPT-4)  & $40 \pm 16$     & $56 \pm 16$     & $52 \pm 16$   & $\mathbf{92 \pm 8}$ \\
                                   & Domain-General Prompt & $44 \pm 16$     & $48 \pm 16$     & $48 \pm 16$   & $\mathbf{72 \pm 16}$ \\
\bottomrule
\end{tabular}
}
\end{table}

\section{Further Prompt Discussion}
\label{appsec:prompts}
\subsection{Extracting Choices}
\label{appsubsec:choice_discussion}

By default, in \Cref{subsec:eval_game24}, we prompt the FM to return a JSON object containing just the numerical index of the choice.
We choose this because of the ease of parsing the response and validating it lies within the correct bounds.
An example prompt is displayed below.

\begin{mybox}{JSON Choice Prompt.}
\small
Reply concisely and exactly with the following JSON format:\\
\{``choice'': X\}\\
where X is the index of the desired choice.
\end{mybox}

When using chain-of-thought as in \Cref{subsec:eval_babyai}, we use the following prompt:

\begin{mybox}{JSON Choice Prompt (Chain of Thought).}
\small
First, briefly reason about your plan.\\
Reply concisely and exactly with the following JSON format:\\
\{``thought'': X, ``choice'': Y\} \\
where X is your reasoning and Y is the index of the desired choice. Make sure Y is a parsable integer.
\end{mybox}

For the TextWorld environment in \Cref{subsec:eval_textworld}, since the action space is much larger, we ask the FM to directly output a text action that we automatically parse.

\begin{mybox}{Decision Making Prompt.}
\small
Please briefly reason about your plan and then output the command in the format `$>$ command'. Ensure only one command is included.
\end{mybox}

We use the regex ``$>$ (.*?)(?:\.|\$)'' (in Perl notation) to parse the command.
We note that the failure rate for both of these options is very low, less than 0.1\% across our evaluation.
Despite this, we include a failsafe that returns a random choice in case of an invalid output.

\subsection{Rejection-based Archive Filtering}
\label{appsubsec:rejection_archive}
The `acceptance-based' archive filter in \Cref{subsec:ige_filter_state} iterates through every new state and asks whether each one is interestingly new and should be added to the archive.
This approach can break down in larger environments where it becomes necessary to explicitly deprecate earlier discoveries that have become irrelevant, in order not to overload the archive (e.g., in \Cref{subsec:eval_textworld}).
In this environment, we use an alternate version of the prompt which first adds all states, and then asks the foundation model to remove the uninteresting states.
An example prompt is shown below.

\begin{mybox}{Rejection-based Archive Filtering Prompt.}
\small
Current state archive: \\
\hl{[State Archive]} \\
Remove outdated states that are no longer relevant to the task, have had all interesting explorations attempted, or have similar states in the archive that show more progress.
\end{mybox}

\subsection{Implementation of State-Conditional Action History}
\label{appsubsec:state_conditional_history}

The state-conditional action history in \ouralgo stores the set of actions previously attempted from each archived state.
For each state in the archive, we maintain a list of actions that have been tried from that state, which helps prevent the foundation model from repeating the same actions and encourages exploration of new actions.
This history is provided in the context when prompting the foundation model for the next action.
Since our state archive is kept at a manageable size (see \Cref{tab:archive_size_analysis}), this additional memory requirement is minimal.

\subsection{Handling Hallucinations during Trajectory Creation.}
Foundation models are known to sometimes produce hallucinations or invalid outputs.
In \ouralgo, the foundation model does not create entire trajectories; instead, it is queried for state/action choices from actual options, and the resulting states are saved into an archive, as in the original Go-Explore.
Therefore, \ouralgo provides robust scaffolding for foundation model agents that explicitly prevent the hallucination of impossible states.
There is a concern that the LLM could pick impossible actions, but we observe that malformed inputs only consist of less than 0.1\% of total actions.
As discussed in \Cref{appsec:prompts}, in these rare cases, we can simply take the simple choice of randomly sampling an action from the environment.
Further work could explore additional safeguards and validation mechanisms to handle hallucinations more effectively.

\section{Hyperparameters}
\label{appsec:hypers}
In this section, we provide the hyperparameters for our empirical evaluation in \Cref{sec:eval}.
We list the hyperparameters for \ouralgo in \Cref{tab:ige_hyp}.
We choose the values for exploratory rollout length based on the average number of steps needed to make `reasonable progress' in the environment.

\setlength{\tabcolsep}{11pt}
\begin{table}[h!]
\centering
\caption{\ouralgo{} Sampling Parameters. TH, TCG, and CC are abbreviations for Treasure Hunter, The Cooking Game, and Coin Collector in TextWorld.}
\begin{tabular}{cccccc}
\toprule
\multirow{2}{*}{\textbf{Hyperparameter}} & \multicolumn{5}{c}{\textbf{Value(s)}} \\
 & \textbf{Game of 24} & \textbf{BabyAI} & \textbf{TH} & \textbf{TCG} & \textbf{CC} \\ 
\midrule
\multicolumn{1}{r}{No.\ state expansions, $N_{\text{state}}$} & 50 & 25 & 24 & 48 & 125  \\
\multicolumn{1}{r}{No.\ exploratory actions, $N_{\text{action}}$} & 3 & 10 & 5 & 5 & 1 \\ 
\bottomrule
\end{tabular}%
\label{tab:ige_hyp}
\end{table}

We list the sampling parameters for GPT-4~\citep{openai2024gpt4} passed via the OpenAI API in \Cref{tab:gpt4_hyp}.

\setlength{\tabcolsep}{17.5pt}
\begin{table}[h!]
\centering
\caption{GPT-4 Sampling Parameters}
\begin{tabular}{rccc}
\toprule
\multicolumn{1}{c}{\multirow{2}{*}{\textbf{Hyperparameter}}} & \multicolumn{3}{c}{\textbf{Value}} \\
\multicolumn{1}{c}{} & \textbf{Game of 24} & \textbf{BabyAI} & \textbf{TextWorld} \\ \midrule
Temperature & 0.7 & 0.7 & 0.3 \\
Max new tokens & 1000 & 1000 & 1000 \\
Response format & JSON Object & JSON Object & Text \\
Version & Turbo-2024-04-09 & o-2024-05-13 & o-2024-05-13 \\ \bottomrule
\end{tabular}%
\label{tab:gpt4_hyp}
\end{table}

We used GPT-4-Turbo for Game of 24 and GPT-4o for BabyAI and TextWorld.
This was purely done to select the version of GPT-4 that was available and the cheapest at the time of running the experiments.
The version of GPT-4 is consistent per environment.
We use a reduced temperature for the TextWorld domain to reduce the possibility of generating malformed responses, as actions are output in free-form natural language.
In our ablations in \Cref{sec:analysis}, we use the `turbo-0125' variant of GPT-3.5.

\subsection{Cost of Experiments}
\label{subsec:cost}

We provide the average cost per task for our algorithm per environment (the number of seeds is specified in \Cref{sec:eval}):

\setlength{\tabcolsep}{68pt}
\begin{table}[h!]
\centering
\caption{Per task API cost for \ouralgo using GPT-4 listed in USD.}
\label{tab:gpt4_cost}
\begin{tabular}{rc}
\toprule
\multicolumn{1}{c}{\textbf{Environment}} & \textbf{API Cost (USD)} \\ \midrule
Game of 24 & 1.04 \\
BabyAI & 2.01 \\
TextWorld & 1.28 \\ \bottomrule
\end{tabular}%
\end{table}

We note that the price per token of the `o-2024-05-13' option is half that of `Turbo-2024-04-09', so we could expect to achieve the same level of results on the Game of 24 with half the price.
The total cost of API access required to perform the final experiments in this paper was under 2,000 USD.
During development, we iterated on \ouralgo with a smaller number of seeds, which represents a small fraction of this cost added on top.

\section{More Related Work and Future Work}
\label{appsec:more_related_work}

\textbf{FM-as-judge.}
We employ FM guidance at all stages of \ouralgo to drive exploration.
FMs as judges~\citep{NEURIPS2023_91f18a12, bradley2023qualitydiversity} have already seen use in decision-making tasks: OMNI~\citep{zhang2024omni} considers FM guidance in multi-task settings to select the most promising next task to train on.
However, focusing on the broader task could miss out on interesting behavior that happens at a more granular level, and thus \ouralgo greatly expands on the integration of FM intelligence into decision-making.
RL from AI Feedback~\citep{bai2022constitutional, lee2024rlaif, klissarov2023motif} considers training RL agents using reward functions derived from FM preferences.
This similarly guides agents towards preferred states but without the intelligence of FMs for action selection.

\textbf{Extension to Stochastic Environments.}
Our current experiments focus on deterministic environments to clearly demonstrate the effectiveness of \ouralgo.
However, \ouralgo can be extended to stochastic environments, similar to how the original Go-Explore~\citep{ecoffet2021goexplore} was extended.
In the 2021 Go-Explore work, a robust goal-conditioned policy is trained via imitation learning on the trajectories found during exploration, enabling generalization to stochasticity.
In the context of \ouralgo, we can similarly collect successful trajectories and provide them in-context to the foundation model, leveraging its ability to generalize and handle stochastic outcomes.
By incorporating past trajectories into the FM's context, \ouralgo can learn to navigate stochastic transitions and maintain robust exploration strategies.
This represents an exciting direction for future research.

\textbf{Relationship with Hierarchical Search and MCTS.}
The Go-Explore framework, and by extension \ouralgo, shares similarities with hierarchical search and Best-First Search (BFS), as it prioritizes exploration from the most promising states.
However, \ouralgo builds on this approach by leveraging the foundation model's intelligence to dynamically assess the interestingness and potential of states, rather than relying on fixed heuristics.
This allows \ouralgo to adaptively explore the search space in a more informed manner.
In extremely hard combinatorial problems like chess and Go, integrating an effective interestingness function is crucial.
If we could train or obtain such a function, combined with a solid value function similar to those used in Monte Carlo Tree Search (MCTS), \ouralgo could potentially discover novel and interesting strategies.
Exploring this integration with MCTS and developing advanced interestingness functions through FMs are promising avenues for future work.

\textbf{Comparison to GFlowNets.}
GFlowNets~\citep{bengio2023gflownetfoundations} aim to sample compositional structures proportionally to a specified reward function, which differs from \ouralgo's goal of exploring and discovering interesting states without predefined rewards.
While both involve generating states through sequential decisions, \ouralgo focuses on leveraging the foundation model's notions of interestingness rather than sampling according to a reward distribution.
Moreover, GFlowNets are not directly intended for hard-exploration or sparse-reward problems.
\ouralgo should therefore be much better in hard exploration tasks where there is little to no reward signal to guide search.
We believe a detailed comparison with GFlowNets is an interesting direction for future research.

\textbf{Comparison with Exploration Methods in Games like Minecraft.}
Recent works in Minecraft, such as Voyager~\citep{wang2023voyager} and Ghost in the Minecraft~\citep{zhu2023ghost}, have employed agents that act in the environment via collections of high-level code or algorithmic policies tailored to Minecraft.
In contrast, \ouralgo provides a generic method to operate in any state and action space, given a minimal description of the environment, by leveraging the intelligence of foundation models at test time.
The code policies used in these Minecraft agents are indeed interesting and could potentially be integrated into an \ouralgo-like framework, representing a promising direction for future research into more efficient exploration agents.
By combining the strengths of both approaches, we could enhance the exploration capabilities of agents in complex environments.

\textbf{Potential Real-life Applications.}
\ouralgo's framework is very general, and we envision that it could be extended to many real-life problems involving exploration in complex spaces.
For example, in synthetic biology, \ouralgo could aid in discovering novel proteins or designing new drugs.
In mathematics, it could assist in exploring mathematical conjectures or finding novel proofs.
Recent work has begun to explore applications of LLM agents across science, such as in biology research~\citep{laurent2024labbenchmeasuringcapabilitieslanguage} and solving olympiad-level math problems~\citep{trinh2024solving}.
On the practical side, LLMs have been adapted for various web-browsing and computer-based tasks~\citep{liu2023agentbench} as useful personal assistants, many of which require exploration across long horizons (e.g., building an app from scratch).
By leveraging \ouralgo, these agents could improve in planning and exploration in such complex tasks.

\textbf{Accelerating Open-endedness Research.}
Open-endedness is often formulated as an unsupervised exploration problem, aiming to develop algorithms that can continually generate novel and diverse behaviors or solutions without predefined objectives.
\ouralgo provides a general mechanism to explore and discover a diverse set of interesting states or solutions in arbitrary environments, much like human scientific exploration.
By leveraging FMs' internalized notions of interestingness, \ouralgo can identify and pursue novel directions that may not be specified explicitly.
This capability could accelerate research in open-endedness by providing a powerful tool for unsupervised exploration in complex domains.

\textbf{Robustness and Potential Biases.}
Foundation models are known to have biases stemming from their training data, which could affect their performance and decision-making in certain environments.
Although representation in pre-training is hard to measure even for open-weight models, we have strong evidence that \ouralgo's success is relatively independent of the base LLM performance.
For instance, in our new results in \Cref{tab:ige_model_comparison}, with Claude Sonnet 3.5 on Game of 24, the naive action selection performance is low (24\%), but \ouralgo raises it significantly to 86\%.
The general question of bias is significant; for example, if a foundation model was biased to think that nothing ``purple'' is interesting, we wouldn't explore states that are ``purple''.
We do not observe such issues in our current evaluation (indeed, we would expect them to manifest less in games and RL/control problems rather than human-centric problems), but this raises an interesting point tied to the wider literature of reducing bias in foundation models in general.
Future work could investigate methods to detect and mitigate potential biases in \ouralgo to ensure fair and unbiased exploration.

\textbf{Potential for Fine-tuning or RL Training.}
Fine-tuning or additional RL training could potentially enhance foundation models' performance in specific tasks.
In the context of \ouralgo, one could envision fine-tuning the foundation model on the trajectories discovered during exploration, further improving its decision-making and ability to generalize.
Our current approach shows that significant gains can already be achieved without additional training, which is advantageous for efficiency.
Additionally, the trajectories discovered by \ouralgo could be used to train traditional RL or imitation learning algorithms, providing valuable data for further improvements.
Exploring these possibilities could lead to more powerful agents capable of solving even more complex tasks.

\textbf{Investigating LLMs' Notions of Interestingness.}
Understanding and formalizing the foundation model's notions of interestingness is an intriguing direction for future research.
Similar to how reward models are collected on human preferences and used to fine-tune LLMs~\citep{ouyang2022traininglanguagemodelsfollow}, we could consider extracting interestingness preferences and fine-tuning an LLM to select more interesting states or actions.
This could then lead to more efficient versions of \ouralgo or even improvements to FMs themselves.
Developing methods to quantify and interpret the interestingness judgments made by FMs could also enhance our understanding of their decision-making processes.

\textbf{Alternate Action Selection Strategies.}
Random action selection and FM action selection represent two extremes in action selection strategies.
Our framework is flexible, and we could consider substituting the action selection policy in \ouralgo with any RL policy.
This would enable us to keep the foundation model \ouralgo scaffolding that prevents detachment and derailment and allow us to use more computationally efficient options to roll out trajectories.
For example, integrating exploration strategies based on intrinsic motivation or curiosity-driven policies could enhance scalability.
Exploring such combinations could provide a balance between performance and computational cost, extending the applicability of \ouralgo to larger and more complex environments.

\end{document}